%% file: ARXIV.tex
\documentclass[11pt]{article}

\usepackage[utf8]{inputenc} 
\usepackage[T1]{fontenc}    
\usepackage{url}            
\usepackage{booktabs}       
\usepackage{amsfonts}       
\usepackage{nicefrac}       
\usepackage{microtype}      
\usepackage{xcolor}         

\usepackage{amsmath, amssymb} 
\usepackage{graphics, graphicx} 
\usepackage[margin=1in]{geometry} 
\usepackage{booktabs} 
\usepackage{units} 
\usepackage{multicol}
\usepackage{bm}
\usepackage{mathtools}
\usepackage{natbib} 
\usepackage{tikz} 
\usetikzlibrary{shapes,decorations}
\usetikzlibrary{arrows.meta}
\usepackage{pgfplots} 
\pgfplotsset{compat=newest} 
\usepackage{comment}
\usepackage{fancyhdr} 
\usepackage[normalem]{ulem} 
\usepackage[T1]{fontenc} 
\usepackage{comment}
\usepackage[shortlabels]{enumitem}
\usepackage[linesnumbered,algoruled,lined]{algorithm2e}
\usepackage[overload]{empheq}
\usepackage{caption}
\usepackage{amsthm}
\usepackage{wrapfig}
\definecolor{mydarkblue}{rgb}{0,0.08,0.45}
\usepackage[colorlinks, citecolor=blue, linkcolor=blue]{hyperref} 
\makeatletter
\newcommand*{\centernot}{%
  \mathpalette\@centernot
}
\def\@centernot#1#2{%
  \mathrel{%
    \rlap{%
      \settowidth\dimen@{$\m@th#1{#2}$}%
      \kern.5\dimen@
      \settowidth\dimen@{$\m@th#1=$}%
      \kern-.5\dimen@
      $\m@th#1\not$%
    }%
    {#2}%
  }%
}
\makeatother

\theoremstyle{plain}
\newtheorem{definition}{Definition}

\newtheorem{assumption}{Assumption}
\newtheorem{theorem}{Theorem}

\newtheorem{proposition}{Proposition}
\newtheorem{remark}{Remark}
\newtheorem{example}{Example}

\newcommand{\stkout}[1]{\ifmmode\text{\sout{\ensuremath{#1}}}\else\sout{#1}\fi}
\usepackage{contour}

\tikzstyle{observed}=[draw, circle, white, text=black, minimum size=6mm]
\tikzstyle{latent}=[draw, circle, blue!75, text=black, minimum size=6mm]
\tikzstyle{deterministic}=[draw, circle, double, blue!75, text=black, minimum size=6mm]
\tikzstyle{noise}=[draw, text=black]

\tikzset{causal/.style={-{Triangle[length=2.5pt, width=3pt]}, line width=1pt},
         exogenous/.style={-{Triangle[length=2.5pt, width=3pt]}, line width=1pt, blue!75},
         difference/.style={-{Triangle[length=2.5pt, width=3pt]}, line width=1pt, red}
}

\newcommand{\indep}{\perp \!\!\! \perp}
\newcommand{\notindep}{\not\!\perp\!\!\!\perp}

\newcommand{\TOv}{{\bf T_O}}

\newcommand{\TLv}{{\bf T_L}}
\newcommand{\bI}{\mathbf{I}}
\newcommand{\bw}{\mathbf{w}}
\newcommand{\bmm}{\mathbf{m}}
\newcommand{\setM}{\mathcal{M}}
\newcommand{\setI}{\mathcal{I}}
\newcommand{\setN}{\mathcal{N}}
\newcommand{\setZ}{\mathcal{Z}}

\newcommand{\new}[1]{{\color{black} #1}}
\newcommand{\warning}[1]{{\color{black} #1}}
\newcommand{\saber}[1]{{\color{black} #1}}
\newcommand{\note}[1]{{\color{black} #1}}

\usepackage{times}
\title{\textbf{Learning Unknown Intervention Targets in Structural Causal Models from Heterogeneous Data}}

\author{
\textbf{Yuqin Yang}\thanks{Equal contribution. Correspondence to: \href{mailto:yuqinyang@gatech.edu}{\texttt{yuqinyang@gatech.edu}}. Accepted at 27th International Conference on Artificial Intelligence and Statistics (AISTATS 2024).} 
\\ Georgia Institute of \\ Technology
\and 
\textbf{Saber Salehkaleybar$^*$} 
\\ Leiden University
\and
\textbf{Negar Kiyavash} 
\\ École Polytechnique \\ Fédérale de Lausanne
}

\date{}

\input{defns}
\begin{document}
\maketitle
\begin{abstract}
We study the problem of identifying the unknown intervention targets in structural causal models where we have access to heterogeneous data collected from multiple environments. The unknown intervention targets are the set of endogenous variables whose corresponding exogenous noises change across the environments. We propose a two-phase approach which in the first phase recovers the exogenous noises corresponding to unknown intervention targets whose distributions have changed across environments. In the second phase, the recovered noises are matched with the corresponding endogenous variables. For the recovery phase, we provide sufficient conditions for learning these exogenous noises up to some component-wise invertible transformation. For the matching phase, under the causal sufficiency assumption, we show that the proposed method uniquely identifies the intervention targets. In the presence of latent confounders, the intervention targets among the observed variables cannot be determined uniquely. We provide a candidate intervention target set which is a superset of the true intervention targets. Our approach improves upon the state of the art as the returned candidate set is always a subset of the target set returned by previous work. Moreover, we do not require restrictive assumptions such as linearity of the causal model or performing invariance tests to learn whether a distribution is changing across environments which could be highly sample inefficient. Our experimental results show the effectiveness of our proposed algorithm in practice.
\end{abstract}
\section{Introduction}
Causal relationships among a set of variables in a system can be modeled by a structural causal model (SCM) where each variable is a function of its direct causes and some exogenous noise. An intervention on a variable can be considered as modifying its causal mechanism, i.e., changing the conditional probability distribution of the intervened variable given its direct causes. In randomized control trials, randomized interventions on a target variable are utilized to estimate the causal effect of the target. 
However, in some applications, we may not have full control in terms of which variables are intervened on. For instance, in recovering 
causal protein-signaling networks from single-cell data \citep{sachs2005causal,ness2017bayesian}, drugs are injected into cells to inhibit or activate some signaling proteins, and gene expression levels are measured. In these experiments, the intervention targets are unknown. Moreover, in some cases, an intervention is done by an unknown source and we must locate the source of the intervention in the system. As an example, microservices systems in cloud clusters are vulnerable to faults such as equipment failures or adversarial attacks. It is crucial to locate the root cause of faulty operation in the system by identifying the source of fault/intervention \citep{aggarwal2021localization,budhathoki2022causal,ikram2022root}. In these examples, the collected data is often heterogeneous and is gathered
from multiple domains/environments where the causal mechanisms of some of the variables are changing across the environments.


In this paper, we consider the problem of learning the unknown intervention targets from a collection of interventional distributions obtained from multiple environments. This problem is closely related to learning an equivalence class of all causal graphs consistent with the collected interventional data. The latter problem has been studied in several work (see the related work in Section \ref{sec:related work}), and some of the proposed methods also provide some information about the locations of intervention targets as a byproduct of the returned equivalence class. These previous methods have several drawbacks such as being limited to linear systems, requiring a huge number of conditional and invariance tests, or lacking the ability to handle latent confounders in the systems 
(see Section \ref{sec:related work} for more details). 


We propose Locating Intervention Target (LIT) algorithm which returns the observed variables that are intervention targets. LIT has two main phases: the recovery phase and the matching phase. In the recovery phase, through a contrastive-learning approach, the exogenous noises corresponding to intervention targets are recovered up to some permutation and component-wise invertible transformation\footnote{In the paper, whenever we say that some exogenous noises can be recovered, it means that they are recovered up to some permutation and component-wise invertible transformation.}. In the matching phase, the recovered exogenous noises are matched to their corresponding observed variables (if any)\footnote{It is noteworthy that a recovered exogenous noise may correspond to a latent variable. In that case, it should not be matched with any observed variable.} by performing conditional independence (CI) tests.
The main contributions of this paper are: 
\begin{itemize}
    \item For the recovery phase, we provide identifiability results for {recovering the exogenous noises whose distributions change across the environments}. In particular, in nonlinear causal models with exogenous noises from an exponential family, the recovery is possible under some mild invertibility assumption (Assumption \ref{assump:tidleg}(a)) for causally sufficient systems (i.e., there are no latent confounders).
    For systems with latent confounders, under some further assumptions (Assumption \ref{assump:tidleg}(b)), we show that the recovery is still possible (Proposition \ref{prop:recover_sources}).
    \item For the matching phase, we prove that LIT algorithm recovers the true intervention targets for causally sufficient systems using the recovered exogenous noises (Theorem \ref{th:main}). LIT algorithm requires merely  quadratic number of CI tests. In comparison, previous work of \citep{jaber2020causal} required an exponential number of CI/invariance tests with respect to number of variables in the system. In the presence of latent confounders, we show that LIT algorithm returns a superset of true intervention targets and present a graphical characterization of the output of recovery (Theorem \ref{th:main_latent}). \new{Unlike previous work, LIT algorithm allows latent confounders to change across environments. Moreover, for the setting studied in the literature (i.e., when latent confounders do not change across environments), our recovery output is more informative than the state-of-the-art (see Remark \ref{remark:comparison} for more details). 
    }
    \item Our experimental results show that LIT outperforms previous work in recovering intervention targets in the presence of latent confounders as well as when the underlying SCM is nonlinear.
\end{itemize}

\section{Preliminaries}
\label{sec:Pre}
In this section, we present the notations used in the paper as well as some necessary background. Upper case letters denote random variables and bold letters indicate sets of random variables. For ease of notation, we also denote the vectorized form of a set of random variables by bold letters. We show the cardinality of set $\Xv$ by $|\Xv|$. We also denote the set $\{1,\cdots,n\}$ by $[n]$.

\textbf{Structural Causal Models.} A structural causal model (SCM) $\mathcal{M}$
is a 4-tuple $\langle \Nv, \Xv, \mathcal{F}, P(\Nv)\rangle$
where $\Nv$ is the set of exogenous noises and $\Xv$ is the
set of endogenous variables. $\mathcal{F}$ represents a collection of functions $\mathcal{F} = \{f_i\}$ such that each endogenous
variable $X_i\in \Xv$ is determined by  $X_i:=f_i(\PA_i,N_i)$ where $\PA_i\subseteq\Xv$ is the set of parents of $X_i$ and $N_i\in\Nv$ is its corresponding exogenous noise. It is assumed that
$\{N_i\}$ are jointly independent. In a given SCM, we may only observe a subset of endogenous variables. Thus, we partition $\Xv$ into two disjoint subsets $\Ov$ and $\Lv$, where $\Ov$ is the set of observed and  $\Lv$ is the set of latent variables. Under the causal sufficiency assumption, we observe all the endogenous variables, i.e., $\Lv=\emptyset$.

The graph $G$ of an SCM is constructed by considering one vertex for each $X_i$ and drawing directed edges from each parent in $\PA_i$ to $X_j$. We assume that the graph $G$ is a directed acyclic graph (DAG), i.e., it contains no directed cycle. We say $X_j$ is an ancestor of $X_i$ if there exists a direct path from $X_j$ to $X_i$. In graph $G$, we denote the set of ancestors and children of $X_i$ by $An_G(X_i)$ and $Ch_G(X_i)$, respectively. We also consider each variable $X_i\in \Xv$ as its own ancestor.  The 
 CI relations can be read from the causal graph using a graphical criterion known as \emph{d-separation} \citep{pearl1988probabilistic}. For disjoint subsets of variables $\Uv, \Vv, \Wv$, we denote the CI relation of $\Uv$ from $\Vv$ given $\Wv$ by $\Uv \indep \Vv |\Wv$. The analogous d-separation statement, $\Uv$ is d-separated from $\Vv$ given $\Wv$ in graph $G$, is written as $(\Uv \indep \Vv | \Wv)_{G}$. 
 In the presence of latent confounders, the causal relationships are often represented by a maximal ancestral graph (MAG). See \citep{richardson2002ancestral} for the definitions of MAGs and inducing paths. 

\textbf{Soft Intervention.} We consider \emph{soft} interventions on a subset of variables such as $\Wv\subseteq \Xv$ of the form obtained by replacing structural assignment $X_i:=f_i(\PA_i,N_i)$ with $X_i:=f_i(\PA_i,N'_i)$ for all $X_i\in \Wv$. $N'_i$ is the new exogenous noise corresponding to $X_i$. Note that in the definition of soft intervention, neither the set of parents nor causal mechanisms $\{f_i\}$ change. In some applications, this operation is more realistic than \emph{hard} interventions, where intervened variables are forced to take a fixed value \citep{varici2022intervention}.
For instance, in molecular biology, the effect of added chemicals to a cell cannot be set to some constant value \citep{eaton2007exact}, or in control theory, for the task of system identification \citep{ljung1998system}, a mathematical model describing the underlying dynamical system is identified by applying certain inputs without changing the dynamics of the system. \warning{Another example is adversarial attacks in cloud systems. A third-party attacker can send corrupted data to the servers in a data center but it might not alter the protocols of communications among them. In this sense, the soft intervention considered is weaker than allowing changes in causal mechanisms inside a system. It is noteworthy that the definition of soft intervention in this paper was also considered in the literature of causal discovery with multiple environments such as in \citep{ghassami2017learning} or in the field of domain adaptation where the causal mechanism is shared across domains \citep{teshima2020few}.} 
\section{Methodology}
\label{sec:method}
\subsection{Problem definition}
\label{subsec:problem}
We consider a multi-environment setting comprised of $D$ environments $\Ecal = \{E_1, ..., E_D \}$. The
underlying causal DAG and the functional mechanism for generating the variables from their parents remain the same across all environments while the distributions of exogenous noises may vary due to some unknown soft interventions. In particular, we have access to a collection of joint distributions over $\Ov$, $\mathcal{P}=\{p_1(\Ov),\cdots,p_D(\Ov)\}$ from $D$ environments. We also denote $p_i(\Nv)$ as the joint distribution over the set of exogenous noises $\Nv$ in environment $E_i$. 
Let $\Tv$ be the set of variables whose exogenous noises are changing across environments, i.e., $\Tv:=\{X_i|\exists d,d'\in [D], p_{d}(N_i)\neq p_{d'}(N_i), 1\leq i \leq n\}$.
These are the variables that are intervened on by some external stimuli and we seek to learn them.
Let $\Nv_{\Tv}:=\{N_i|X_i\in \Tv\}$ be the set of exogenous noises whose distributions are changing across the environments,
and $\TOv=\Tv \cap \Ov$ be the set of observed variables that are intervened on. Similarly, denote the set of intervention targets in the latent part by $\TLv = \Tv \cap \Lv$.
Note that under causal sufficiency,  $\TOv=\Tv$. 
Our goal is to locate interventions, i.e., recover the unknown observable targets of interventions $\TOv$ from merely the observational distributions $\mathcal{P}$ over the multiple environments. 


In the following, we present our method for learning the intervention targets, which has two main phases: the recovery phase and the matching phase. In Section \ref{section:recovery}, we present the recovery phase, which is to recover the set of exogenous noises whose distributions are changing across the environments (up to some permutation and component-wise invertible transformations). Next, we present the matching phase in Section \ref{section:matching}, where we match the recovered noises with the corresponding variables in $\Xv$ in order to learn $\TOv$. 

\subsection{Recovery Phase} \label{section:recovery}
For a given SCM $\mathcal{M}$, due to the assumption that the causal graph is a DAG, each observed variable $X_i\in \Xv$ can be written as $X_i=g_i(\Nv)$ where function $g_i$ only depends on exogenous noises corresponding to the ancestors of $X_i$. We collect all these equations in the vector form $\Xv = \gv_{\mathcal{M}} (\Nv)$ where $\gv_{\mathcal{M}}:\mathbb{R}^{n}\rightarrow\mathbb{R}^n$ and $n=|\Xv|$ is the number of variables in the system. We call function $\gv_{\mathcal{M}}$, the ``mixing function'' of SCM $\mathcal{M}$.

\warning{As we have access to a collection of distributions in $\mathcal{P}$, we will exploit the heterogeneity in data to recover the exogenous noises in $\Nv_{\Tv}$. Specifically, we utilize a contrastive learning approach. We observe auxiliary variable $U$ indicating the index of the environment, and train a nonlinear regression model with universal approximation capability\footnote{Universal
approximation capability refers to the ability to approximate any Borel measurable function to any desired degree of accuracy. See \citep{hornik1989multilayer} for more detail.} for the supervised learning task (Please refer to Appendix \ref{app:recovery} for more details). We consider an exponential family for the distributions of the exogenous noises (see Appendix \ref{app:recovery} for the definition). To ensure that the noises in $\NT$ can be recovered, we require the following assumption:}

\begin{assumption}
For a given SCM $\mathcal{M}$, we assume that either: (a) the corresponding mixing function $\gv_{\mathcal{M}}$ is invertible, or (b) there exists an invertible function $\tilde{\gv}:\mathbb{R}^{|\Ov|}\rightarrow \mathbb{R}^{|\Ov|}$ such that $\tilde{\gv}(\Ov)=(\Nv_{\Tv}, \Vv)$ where $\Vv\in \mathbb{R}^{|\Ov|-|\Tv|}$ is a random vector satisfying $\Vv\indep U$ and $\Nv_{\Tv}\indep \Vv|U$.
\label{assump:tidleg}
\end{assumption}

\warning{Assumption \ref{assump:tidleg}(a) is standard in contrastive learning algorithms under the causal sufficiency assumption. It is satisfied for all acyclic linear SCMs and nonlinear additive noise models. To extend the results to the latent confounder setting, we added Assumption \ref{assump:tidleg}(b). In particular, vector $\Vv$ is the recovered part that is invariant across the environments and is independent of $\Nv_{\Tv}$ given the index of the environment. It corresponds to a function of exogenous noises whose distributions do not change across the environments.}

\begin{proposition}
    Assume that $\min(D-1,|\Ov|)\geq|\Tv|$ (Recall that $D$ is the number of environments and $\Ov$ and $\Lv$ are the set of observed and latent variables in the system, respectively).
       By utilizing the contrastive-learning approach, the exogenous noises in $\Nv_{\Tv}$ can be recovered up to some permutation and component-wise strictly monotonic transformations with measure one in the following two settings: 1)  $\Lv=\emptyset$ under Assumption \ref{assump:tidleg}(a). 2) $\Lv\neq\emptyset$ under Assumption \ref{assump:tidleg}(b).

    \label{prop:recover_sources}
\end{proposition}

Please refer to Appendix \ref{app:recovery} for a detailed description of the contrastive learning approach and extra discussions about when Assumption \ref{assump:tidleg} is satisfied.

\begin{remark}
Note that the assumption on the number of environments $D$ is required for recovering the noises in $\mathbf{N}_{\mathbf{T}}$ without any prior knowledge of $f_i$'s functional form. We can relax the assumptions on the number of environments by adding restrictions on $f_i$ (or equivalently on $\mathbf{g}_{\mathcal{M}}$). \note{For example, for piecewise linear 
$\mathbf{g}_{\mathcal{M}}$, under additional assumptions on the exogenous noises,} we can recover $\mathbf{N}_{\mathbf{T}}$
from merely two environments in causally sufficient cases, based on the results in \citep{kivva2022identifiability}. 
\end{remark}

\subsection{Matching Phase} \label{section:matching}
\warning{Throughout the matching phase, we assume that the exogenous noises in $\Nv_{\Tv}$ are recovered up to some permutation and component-wise invertible transformations. Denote the recovered noise corresponding to the noise $N_i\in \NT$ as $\tilde{N}_i$ (i.e., $\tilde{N}_i$ is an invertible transformation of $N_i$), and denote the collection of all the recovered noises as $\TNT$. Note that we cannot learn the correspondence of the recovered noises to the true noises due to permutation indeterminacy according to Proposition \ref{prop:recover_sources}. However, since a variable $X_i$ is in $\Tv$ if and only if its exogenous noise $N_i\in \NT$, we can recover the intervention targets by matching the observed variables to the recovered noises.\footnote{Due to the permutation indeterminacy, there exists a one-to-one mapping $\sigma$ that maps each noise in $\NT$ to a distinct noise in $\TNT$. For notation simplicity, we denote $\sigma(N_i)\in\TNT$ as $\tilde{N}_i$ for each $N_i\in \NT$. 
The matching phase aims to match each recovered noise to its corresponding observed variable (if such a variable exists). Note that a recovered noise may correspond to a latent confounder. Further, an observed variable does not correspond to any recovered noise if it is not in the intervention target set.
}

In the rest of this section, we will show how to use $\TNT$ from the recovery phase to learn the intervention targets. In particular, in Section \ref{section:t-faithfulness}, we define a new notion of faithfulness called T-faithfulness based on an \emph{augmented graph}. In Section \ref{section:matching_sufficiency}, we study causally sufficient models. We present the conditions and the algorithm for recovering the intervention targets, and show that the intervention target set $\Tv$ can be uniquely identified with quadratic number of CI tests. 
In Section \ref{section:matching_latent}, we study the model with latent confounders. We show that by adding and changing some of the conditions in the causally sufficient case, the intervention targets $\TOv$ can be identified up to its superset called \emph{candidate intervention target set}, which is defined through an \emph{auxiliary graph} (see Definition \ref{def:auxillary_graph}) constructed from the true model.
}

\subsubsection{T-faithfulness assumption}\label{section:t-faithfulness}
For a given SCM $\mathcal{M}$ with the causal graph $G$ and intervention targets $\Tv$, we construct an augmented graph $G_{\Tv}$ as follows. For each variable $X_i\in \TOv$ with corresponding exogenous noise $N_{i}$ (recall that $\Tv_{\Ov}$ and $\Tv_{\Lv}$ are the sets of intervention targets in the observed and latent variables, respectively), we add vertex $N_{i}$ and edge $N_{i}\rightarrow X_i$ to $G_{\Tv}$. Further, for each latent confounder $X_l\in \TLv$ with corresponding recovered noise $N_l$, we replace $X_l$ with $N_l$ since $X_l$ can be recovered up to an invertible transformation. We denote the set of noises corresponding to the variables in $\Tv_{\Lv}$ by $\Nv_{\TLv}$. Following this construction, variables in $G_\Tv$ consist of all changing noises in $\Nv_\Tv$, and all variables in $\Xv\setminus \TLv$.


It can be shown that the joint distribution $p(\Xv\setminus \TLv,\Nv_{\Tv})$ satisfies Markov property with respect to graph $G_{\Tv}$ (see Appendix \ref{proof: Markov property}). However, in order to infer the graphical properties of the augmented graph from only observed variables $\Ov$ and the recovered noises $\TNT$, we need a form of faithfulness.
 \begin{assumption}[T-faithfulness]
     \warning{

      The model is T-faithful to the augmented graph $G_{\Tv}$, in the sense that for any noise $N_i\in \Nv_\Tv$, observed variable $X_k\in \Ov$, and disjoint sets $\Wv_1\subseteq \Ov\setminus \{X_k\}$, $\Wv_2\subseteq \Nv_\Tv\setminus\{N_i\}$:
     $(N_i\indep X_k | \Wv_1, \Wv_2)_{G_\Tv}$ if and only if $\tilde{N}_i\indep X_k | \Wv_1, \tilde{\Wv}_2$, where $\tilde{N}_l$ is an arbitrary invertible transformation of $N_l$ for any $N_l\in \NT$, and $\tilde{\Wv}_2=\{\tilde{N}_j|N_j\in \Wv_2\}$.
     
     }
     
     \label{assump:faithfulness}
\end{assumption}

Assumption \ref{assump:faithfulness} implies that for any changing exogenous noise $N_i\in \Nv_\Tv$ and observed variable $X_k$, the recovered noise $\tilde{N}_i$ is (marginally or conditionally) dependent on $X_k$ if and only if $N_i$ and $X_k$ are d-connected
in $G_{\Tv}$. Therefore, given observed variables and the recovered noises, we can construct the indicator set $\Iv_i:=\{\tilde{N}_{j}\in \TNT|\tilde{N}_{j} \notindep X_i \}$ for each variable $X_i$ in $\Ov$, which is the set of recovered noises that are dependent on $X_i$. Under Assumption \ref{assump:faithfulness}, the indicator set $\Iv_i$ corresponds to all the noises in $\NT$ that are ancestors of $X_i$ in $G_{\Tv}$, i.e., $An_{\GT}(X_i)\cap \NT$. 
Define $\mathcal{I}$ as the collection of sets $\{\Iv_i | i\in [n]\}$.
In the following, we show how to identify the intervention targets by matching the recovered noises with the observed variables, based on the indicator sets and limited number of extra CI tests.

\subsubsection{Matching phase under causal sufficiency}\label{section:matching_sufficiency}
For each variable $X_i$, define the \emph{possible parent set} $\Sv_i$ as the set of variables whose indicator set is a strict subset of $\Iv_i$, i.e., $\Sv_i:=\{X_j|\Iv_j\subsetneq \Iv_i,1\leq j\leq n\}$. Define the \emph{residual set} $\setN_i$ as the set of noises in $\Iv_i$ that do not belong to any indicator set of the variables in $\Sv_i$, i.e. $\bI_i\setminus \cup_{j:X_j\in \Sv_i} \Iv_j$. $\Sv_i$ includes a subset of the ancestors of $X_i$, and no descendants of $X_i$ are included in $\Sv_i$. Further, $\setN_i$ represent the recovered noises in $X_i$ that do not affect $X_i$ through variables in $\Sv_i$. Under the causal sufficiency assumption, $|\setN_i|$ is either 0 or 1 (see Appendix \ref{app:proof of prop 3}). The following proposition provides the conditions for checking whether a variable $X_i$ is in the intervention target set or not given the indicator sets $\setI$ and $\Sv_i$. Equipped with this proposition, we devise LIT Algorithm (Algorithm \ref{alg1}) which recovers $\Tv$ under the causal sufficiency assumption.

\begin{proposition}
    Under causal sufficiency and  Assumption \ref{assump:faithfulness}, \warning{for each variable $X_i$,} the following statements hold:
    \begin{enumerate}[(I), leftmargin=*]
        \item \label{proposition-1}
        $X_i\not\in\Tv$ if the residual set \warning{$\setN_i = \emptyset$}.
        \item 
        \label{proposition-2}
        $X_i\in\Tv$ if $\setN_i\neq \emptyset$ and $\Iv_i$ is unique in $\mathcal{I}$.
        \item \warning{If $\setN_i\neq \emptyset$ and $\Iv_i$ is not unique,} let $\setX_i=\{X_{i_1},\cdots,X_{i_p}\}$, $p\geq 2$, be the set of all variables with the same indicator set as $X_i$, including $X_i$ itself.\footnote{As all the variables in $\setX_i$ have the same indicator set, \warning{their corresponding possible parent sets $\Sv_{i_k}$ and residual sets $\setN_{i_k}$} are also equal. In the statement of the proposition, we use $\Iv_{i}$, $\Sv_{i}$, $\setN_{i}$ to denote the \warning{indicator set, possible parent set and residual set} corresponding to any variable in $\{X_{i_1},\cdots,X_{i_p}\}$.} 
        \warning{Suppose $\setN_{i}=\{\tilde{N}_l\}$ for some $\tilde{N}_l\in \TNT$}. Then, the variable $X_{i_k}$ satisfying the following condition is the only variable from $\setX_i$ that is in $\Tv$\warning{, i.e., $X_{i_k}\in \Tv$ if and only if all other variables in $\setX_i$ are independent of $\tilde{N}_l$ conditioned on $X_{i_k}$ and $\Sv_{i}$.} 
        \label{proposition-3}
        \begin{equation}
        \begin{split}
              \tilde{N}_l \indep X_{i_j} | \{X_{i_k}\}\cup \Sv_{i}, \quad\forall j \in [p]\setminus \{k\}.
              \label{eq:third_cond}
        \end{split}
        \tag{C1}
        \end{equation}
    \end{enumerate}
\label{th:main_condition}
\end{proposition}
\warning{Recall that an observed variable is in the intervention target set $\Tv$ if and only if its corresponding exogenous noise is recovered in $\TNT$.} The statement in \ref{proposition-1} holds because if $X_i\in\Tv$, then $\tilde{N}_i$ cannot appear in $\Iv_j$ for any non-descendant $X_j$ of $X_i$. When $\setN_i\neq \emptyset$, this means that the only noise \warning{$\tilde{N}_l \in \setN_i$} is either the exogenous noise of $X_i$, or the exogenous noise of some ancestor of $X_i$ whose indicator set is the same as $\Iv_i$. We can then use conditions \ref{proposition-2} and \ref{proposition-3} to \warning{further distinguish between these two cases}. If there are no other variables with the same indicator set (i.e., $\Iv_i$ is unique), then $X_i\in \Tv$. Otherwise, among the variables with the same indicator set, 
there is only one variable $X_{i_k}$ that \warning{corresponds to} $\tilde{N}_l$ and belongs to the intervention target set, and the rest are descendants of $X_{i_k}$. Herein, we use \eqref{eq:third_cond} to find such $X_{i_k}$, as given $X_{i_k}$ and $\Sv_{i}$, $\tilde{N}_l$ becomes independent from $X_{i_j}$ for all $j\neq k$ under T-faithfulness assumption.
Please refer to Appendix \ref{app:example} for an example explaining how the aforementioned three conditions can be used to recover $\Tv$.

\begin{wrapfigure}{r}{0.5\textwidth}
\hfill
\begin{minipage}{0.49\textwidth}
\begin{algorithm}[H]
\SetAlgoVlined
\LinesNumbered
\DontPrintSemicolon
\caption{LIT algorithm}\label{alg1}
Obtain $\TNT$ and $\setI$; $\Uv \gets \Xv$; $\Kv\gets \emptyset$;\;
\For{$X_i \in \Xv$}{
\lIf{ 
\warning{$\setN_i = \emptyset$}
}{$\Uv \gets \Uv \backslash \{X_i\}$;} \label{line:cond1}
\lElseIf(\tcp*[h]{only with latent confounder}){\eqref{eq:first_cond_latent} holds}{$\Uv \gets \Uv \backslash \{X_i\}$;}
\lElseIf{$\Iv_i$ is unique}{$\Kv\gets \Kv \cup \{X_i\}$, $\Uv \gets \Uv \backslash \{X_i\}$;}
}\label{line:cond2}
Partition $\Uv$ to disjoint subsets $\Uv_1,\cdots,\Uv_r$ according to the indicator sets;\; \label{line:partition}
\For{$\Uv_i \in \{\Uv_1,\cdots,\Uv_r\}$}{
 {$\Kv_{\Uv_i}\gets$ Find $X_{i_k}\in \Uv_i$ satisfying \eqref{eq:third_cond} under causal sufficiency (resp. remove the subset of variables in $\Uv_i$ satisfying \eqref{eq:third_cond_latent} in the presence of latent confounders);\;}
$\Kv\gets \Kv\cup \Kv_{\Uv_i}$; \label{line:cond3}
}
\Return{\Kv}
\end{algorithm}
\end{minipage}
\end{wrapfigure}
Based on Proposition \ref{th:main_condition}, we propose LIT algorithm (see Algorithm \ref{alg1}) which returns a \emph{candidate intervention set} $\Kv$. 
Specifically, we first check if a variable can be added to or excluded from $\Kv$ according to \ref{proposition-1} and \ref{proposition-2}. We then partition the remaining variables in $\Uv$ into disjoint subsets (which correspond to the collection of all $\setX_i$ in \ref{proposition-3}), and find the candidate in each set (denoted by $\Kv_{\Uv_i}$) using condition \ref{proposition-3}. \warning{Note that LIT algorithm only requires quadratic number of CI tests: $O(n|\Tv|)$ for constructing the indicator set, and at most $O(n^2|\Tv|)$ for checking \eqref{eq:third_cond}. This is a significant reduction from the exponential number of independence/invariance tests with respect to $n$  in the literature  \citep{jaber2020causal, mooij2020joint}.}



\begin{theorem}
    Under causal sufficiency and T-faithfulness assumption (Assumption \ref{assump:faithfulness}), Algorithm \ref{alg1} uniquely identifies 
    the intervention target set $\Tv$, i.e., $\Kv=\Tv$.
    \label{th:main}
\end{theorem}

\begin{remark}
We show that there is a connection between our approach here for learning the intervention targets and an existing algorithm for the task of causal discovery in linear SCMs with deterministic relations \citep{yang2022causal}. This allows us to design a more efficient algorithm for recovering the set $\Tv$. Specifically,
conditions \ref{proposition-1}-\ref{proposition-3} can be checked along with the partitioning of the sets $\Uv$ (line 6) without computing $\Sv_i$ for every $X_i$.
Please see Appendix \ref{app:efficient alg} for more details.
\label{remark:algorithm}
\end{remark}
\subsubsection{Matching phase in the presence of latent confounders} \label{section:matching_latent}
We extend our results from causally sufficient case to the case where latent confounders are present. Unlike previous work \citep{jaber2020causal, varici2022intervention}, we allow latent confounders to be in $\Tv$. 
In this case, an exogenous noise in $\Nv_\Tv$ may correspond to either an observed variable or a latent confounder, and the task is to recover the observed variables that are in the intervention target set, i.e., $\TOv$. 
Unlike the causally sufficient case, $\TOv$ is not always uniquely identifiable. \warning{However, by modifying the LIT algorithm according to the conditions in Proposition \ref{th:condition_latent} below, the algorithm can recover a superset $\Kv$ of $\TOv$. 
Proposition \ref{th:condition_latent} provides conditions for finding variables that do not belong to $\TOv$. In particular, compared with Proposition \ref{th:main_condition}, the statement in \ref{proposition-1} still holds, while condition \ref{proposition-3} is replaced by condition \ref{proposition-3-L}. Further, the statement in \ref{proposition-2} does not hold anymore, and we have one extra condition \ref{proposition-1-L} for excluding variables from $\TOv$.
%


\begin{proposition}
\label{th:condition_latent}
In the presence of latent confounders, under Assumption \ref{assump:faithfulness}, for each observed variable $X_i$, the following statements hold:
\begin{enumerate}[(I)]
    \item 
    $X_i\not\in\TOv$ if the residual set $\setN_i = \emptyset$.
    \setcounter{enumi}{3} 
    \item  \label{proposition-1-L}
    $X_i\not\in\TOv$ if $\setN_i \neq \emptyset$ and every recovered noise in the residual set $\setN_i$ belongs to at least one other indicator set $\Iv_{j_l}$ for some $X_{j_l}\in \Ov$, where $\Iv_{j_l}$ is not a superset of $\Iv_i$:
    \begin{equation}
    \forall \tilde{N}_l\in \setN_i, ~\exists j_l \text{ s.t. } \tilde{N}_l\in \Iv_{j_l}, \text{ and } \Iv_i\not\subseteq \Iv_{j_l}.
    \tag{A}
    \label{eq:first_cond_latent}
    \end{equation}
    
    \renewcommand{\labelenumi}{(III-L)}
    \item If $\setN_i \neq \emptyset$ and condition \eqref{eq:first_cond_latent} does not hold, let $\mathcal{X}_{i} = \{X_{i_1},\cdots,X_{i_p}\}$ be the set of all variables with the same indicator set as $X_i$, including $X_i$ itself. Then for each $j\in [p]$, $X_{i_j}\not\in\TOv$ if it is independent of some recovered noise $\tilde{N}_l$ in $\setN_i$, conditioned on certain subsets of observed variables in $\setX_i$, $\Sv_i$ and all other recovered noises in $\Iv_i$:
    \begin{equation}
    \begin{aligned}
        \exists K \subseteq [p]\setminus \{j\},~ \exists\Sv \subseteq \Sv_{i},~ \exists \tilde{N}_l\in \setN_i \text{ s.t. }
        \tilde{N}_l \indep X_{i_{j}} | \left(\cup_{k'\in K} X_{i_{k'}}\right) \cup \Sv \cup \left( \Iv_i\setminus \{\tilde{N}_{l}\}\right).
    \end{aligned}
    \label{eq:third_cond_latent}
    \tag{C2}
    \end{equation} 
    \label{proposition-3-L}
\end{enumerate}
\end{proposition}}


The statement in \ref{proposition-1-L} holds because if $X_i\in \TOv$, then its corresponding exogenous noise $\tilde{N}_i$ must be in $\setN_i$. Any variable (such as $X_{j_l}$) that is dependent on $\tilde{N}_i$ must be a descendant of $X_i$, and hence have $\Iv_i\subseteq \Iv_{j_l}$.
For \ref{proposition-3-L}, similar to the 
argument for condition \ref{proposition-3},
there is at most one variable (say $X_{i_k}$) in $\setX_i$ that belongs to $\TOv$. Moreover, if $X_{i_k}\in \TOv$, then all other variables in $\setX_i$ are its descendants.
The recovered noises in $\mathcal{N}_i$ cannot be conditionally independent of $X_{i_k}$, as they correspond to either $X_{i_k}$ or some latent confounder that is a parent of $X_{i_k}$. 
Therefore,
if an observed variable $X_{i_j}\in \mathcal{X}_{i}
$ is conditionally independent of a recovered noise in $\mathcal{N}_i$ given some other variables in the system, then it cannot be an intervention target. 
Note that under the causal sufficiency assumption, $X_{i_k}$ and $\Sv_{i}$ are sufficient for the conditioning set. Therefore condition \ref{proposition-3-L} reduces to \ref{proposition-3}. When latent confounders are present, $X_{i_k}$ and $\Sv_{i}$ may not be sufficient. However, condition \ref{proposition-3-L} states that in order to perform such a CI test, it suffices to consider subsets of $
\mathcal{X}_{i}
\backslash\{X_{i_j}\}$ and $\Sv_{i}$ in the conditioning set as their union contains all the ancestors of $X_{i_j}$ among the observed variables. 


\warning{Based on Proposition \ref{th:condition_latent}, we update the LIT algorithm in the presence of latent confounders. We check condition \ref{proposition-1-L} in line 4, and replace condition \ref{proposition-3} by \ref{proposition-3-L} in line 8. 
We keep line 5 in the latent case. In fact, if $X_i$ is not ruled out by conditions \ref{proposition-1} and \ref{proposition-1-L}, it is added to $\Kv$ if $\Iv_i$ is unique as it cannot be ruled out by condition \ref{proposition-3-L} either.
However, the uniqueness of the indicator set does not necessarily imply that the variable belongs to $\TOv$. Lastly, note that under the causal sufficiency assumption, condition \ref{proposition-1-L} is automatically satisfied, and condition \ref{proposition-3-L} reduces to condition \ref{proposition-3}. Hence the algorithm remains consistent with the causally sufficient case.}

\begin{figure*}[t]
    \centering
    \includegraphics[width=0.95\textwidth]{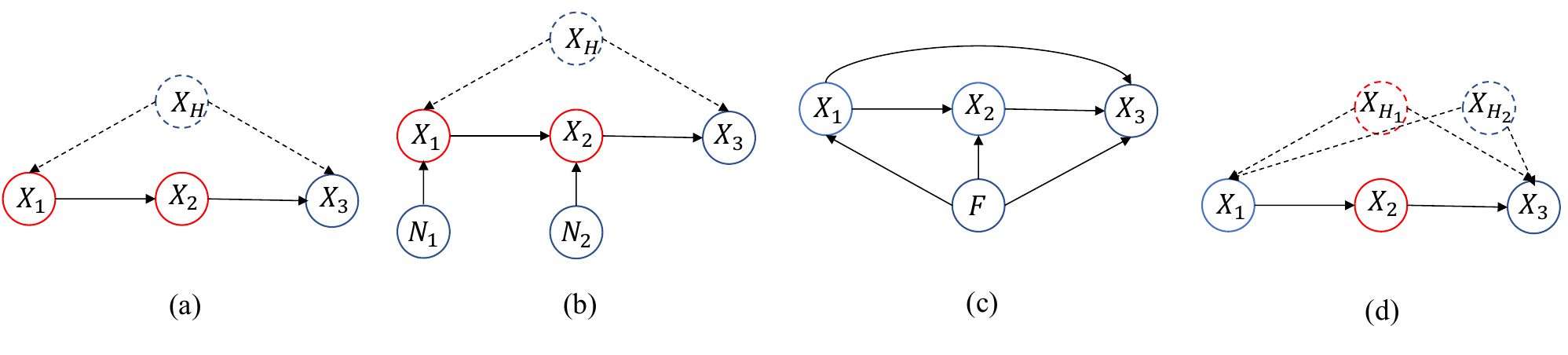}
    \caption{(a) The causal graph of the SCM considered in Example \ref{example:Th2}. (b) The corresponding auxiliary graph according to Definition \ref{def:auxillary_graph}. (c) The MAG of the augmented graph defined in \citep{jaber2020causal}, which indicates the output of their algorithm. (d) The causal graph of an alternative SCM that has the same auxiliary graph.}
    \label{fig:example_2}
\end{figure*}

\begin{example}
Consider an SCM whose corresponding causal graph is depicted in Figure \ref{fig:example_2}(a). It includes three observed variables ${X_1, X_2, X_3}$ and a latent confounder $X_H$, where $\Tv=\{X_1,X_2\}$ (shown in red). Suppose we recovered two noises $\tilde{N}_1$, $\tilde{N}_2$ that correspond to $X_1$, $X_2$, respectively. We have $\Iv_1=\{\tilde{N}_1\}$ and $\Iv_2= \Iv_3=\{\tilde{N}_1, \tilde{N}_2\}$. Following the LIT algorithm, we find that $\Iv_1$ is unique and conditions \ref{proposition-1} and \ref{proposition-1-L} do not hold for $X_1$.
Therefore $X_1\in \Kv$. Further, $\tilde{N}_2 \indep X_3 | X_1, X_2$, and $\tilde{N}_2 \not\indep X_2 | \Sv$ for all $\Sv\in \{\emptyset, \{X_1\}, \{X_3\}, \{X_1, X_3\}\}$. Therefore $X_2\in \Kv$, $X_3\not\in \Kv$ according to condition \ref{proposition-3-L}. In conclusion, we have $\Kv=\{X_1, X_2\}$. Please note that $\Kv$ could be a strict superset of $\Tv_{\Ov}$ in some cases (see Example \ref{example:Th2}).
\label{example: latent1}
\end{example}



In the following, we provide a theoretical analysis of the candidate intervention target set $\Kv$ returned by LIT algorithm in the presence of latent confounders. In particular, we show that $\Kv$ contains the true intervention targets in the observed variables (i.e., $\TOv$). Further, we provide a graphical characterization of what other types of variables are also included in the set $\Kv$, using the notion of \emph{auxiliary graph} which is defined as follows.

\begin{definition}[Auxiliary graph]
\label{def:auxillary_graph}
For each observed variable $X_i$, denote $\Iv_0(X_i)$ as $An_{\GT}(X_i)\cap \NT$.
Given an SCM and its corresponding augmented graph $\GT$, the auxiliary graph $Aux(G_\Tv)$ is constructed from $\GT$ as follows:
\begin{enumerate}[(a), leftmargin=*]
\item For each $X_i\in \TOv$ with its corresponding exogenous noise $N_{i}$, add the edge $N_{i} \rightarrow X_j$ if (i) there is an inducing path between them relative to $\Lv \setminus \TLv$ in $\GT$ 
(i.e., there is an edge between $N_i$ and $X_j$ in the MAG corresponding to $\GT$)
, and (ii) $\Iv_0(X_i)=\Iv_0(X_j)$.
\item (i) For each $\tilde{N}_l\in \TNTL$ (noise that corresponds to a variable in $\TLv$) and each of its child $X_i$, keep the edge $\tilde{N}_l \rightarrow X_i$ if for any other child $X_j$ of $\tilde{N}_l$ in $G_\Tv$, 
\warning{$\Iv_0(X_i)\subseteq\Iv_0(X_j)$}. Otherwise remove the edge $\tilde{N}_l \rightarrow X_i$. (ii) For any remaining edge $\tilde{N}_l \rightarrow X_i$, add (remove) the edge $\tilde{N}_{l} \rightarrow X_k$ if there is an (no) inducing path between $X_k$ and all (some) parents of $X_i$ in $\TNTL$ relative to $\Lv \setminus \TLv$ in $G_\Tv$, and 
$\Iv_0(X_k)=\Iv_0(X_i)$.
\end{enumerate}
\end{definition}

Recall from Section \ref{section:t-faithfulness} that under Assumption \ref{assump:faithfulness}, $An_{\GT}(X_i)\cap \NT$ can be recovered by $\Iv_i$. Therefore we use $\Iv_0(X_i)$ to represent the true value of $\Iv_i$ which does not depend on the recovery output. Further, it is noteworthy that while part (b)(ii) in Definition \ref{def:auxillary_graph} includes both adding and removing of the edges, the addition or removal of edges in the step is independent of the order of the edge selection $\tilde{N}_l\rightarrow X_i$.





\begin{theorem}
In the presence of latent confounders and Assumption \ref{assump:faithfulness}, the candidate intervention target set $\Kv$ returned by LIT algorithm is the set of observed variables that are children of $\NT$ in $Aux(\GT)$, i.e., $\Kv = \cup_{N_i\in \NT} Ch_{Aux(\GT)}(N_i)$.
\label{th:main_latent}
\end{theorem}

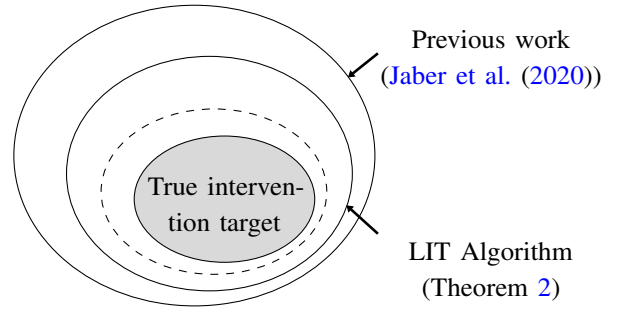
\begin{wrapfigure}{r}{0.5\textwidth}
\hfill
\begin{minipage}{0.49\textwidth}
\hfill
\begin{tikzpicture}[scale=.2]
\centering

\draw (-3.2,.2) ellipse (12cm and 10cm);
\node[text width=3.5cm, text centered] at (16.5, 6.6) {\small Previous work \\ (\cite{jaber2020causal}) }; 
\node[text width=3.5cm, text centered] at (-7.5, 6.6) { };

\path[causal] (9, 7) edge node {} (7., 5.4);

\draw (-2.2,-1) ellipse (9.5cm and 7.8cm); 

\node[text width=3.5cm, text centered] at (16.5, -7.5) {\small LIT Algorithm\\ (Theorem \ref{th:main_latent})}; 

\path[causal] (9, -5) edge node {} (6.9, -3.1);

\draw[dashed] (-1.9,-2.2) ellipse (7.5cm and 5.5cm);

\draw[fill=gray!30] (-1.2,-2.7) ellipse (6cm and 4.2cm); 
\node[text width=4cm, text centered] at (-1.2,-3.2) {\small True interven-\\tion target};

\end{tikzpicture} 
\caption{Comparison of the recovery outputs when latent confounders are not in $\Tv$. The dashed circle represents the theoretical limitation of the recovery of $\Tv$. It may include observed variables that are theoretically indistinguishable from the true intervention targets due to the presence of latent confounders.}
\label{fig:summary}
\end{minipage}
\end{wrapfigure}

Theorem \ref{th:main_latent} gives a graphical characterization of the recovered candidate intervention set $\Kv$.
In particular, according to Definition \ref{def:auxillary_graph}, $\TOv$ is a subset of $\Kv$. This is because the edge from $X_i\in \TOv$ to its corresponding exogenous noise $N_{i}$ in $\GT$ is not removed. This means that LIT algorithm returns a superset of $\TOv$. Further, two other types of variables are added to $\Kv$ according to the conditions in part (a) and part (b) of Definition \ref{def:auxillary_graph}, respectively. 


\begin{remark}
\label{remark:comparison} We can draw the following observations from Theorem \ref{th:main_latent}.
First, under the causal sufficiency assumption, Theorem \ref{th:main_latent} implies that $\TOv$ can be uniquely identified. This is because no edges are added to $Aux(\GT)$ compared with $\GT$ according to Definition \ref{def:auxillary_graph}. 
Second, if all latent variables are not intervention targets (i.e., $\Tv=\TOv$), then our identifiability result is stronger than the existing results in \citep{jaber2020causal,varici2022intervention}. In particular, \cite{jaber2020causal} and \cite{varici2022intervention} showed that in this case, $\TOv$ can only be identified up to the neighbors of $\Nv_{\Tv}$ in the MAG corresponding to $\GT$.
We improve their results by adding part (a)(ii) in Definition \ref{def:auxillary_graph}, due to the recovery of $\tilde{\Nv}_{\Tv}$.  
See \note{Figure \ref{fig:summary} for a diagram comparing the recovery outputs} and Example \ref{example:Th2} below. 
\end{remark}
\begin{example}
    Consider the same example as in Example \ref{example: latent1}. Following the results in \citep{jaber2020causal,varici2022intervention}, the MAG of the augmented graph defined in \citep{jaber2020causal} is shown in Figure \ref{fig:example_2}(c). The recovery output of their algorithms is $\{X_1, X_2, X_3\}$ as all the observed variables are the neighbor of the $F$-node defined in their work.
    On the contrary, LIT algorithm has a more accurate recovery of the intervention targets. Specifically, 
    the auxiliary graph is shown in Figure \ref{fig:example_2}(b). The edge from $N_1$ to $X_3$ is not added since $\Iv_1 \neq \Iv_3$ (which violates part (a)(ii) in Definition \ref{def:auxillary_graph}), and the edge from $N_2$ to $X_3$ is not added because there is no inducing path (which violates part (a)(i)). Therefore $\Kv=\{X_1, X_2\}$, which is the same as the output in Example \ref{example: latent1}.
    Lastly, note that $\Kv$ is not always equal to $\TOv$. Consider the causal graph in Figure \ref{fig:example_2}(d) where $\Tv=\{X_{H_1},X_2\}$.
    Its corresponding auxiliary graph is exactly the one in Figure \ref{fig:example_2}(b). However, 
    $\TOv=\{X_2\}\subsetneq \{X_1,X_2\}=\Kv$, and we cannot distinguish whether $X_1$ or $X_{H_1}$ is the intervention target.   
\label{example:Th2}
\end{example}

\begin{remark}
We note that LIT algorithm is capable of identifying intervention targets for each environment through extra invariance tests. In particular, for each pair of environments, we can conduct an invariance test on the samples of recovered noise in two environments.
\end{remark}
\begin{figure}[t]
\centering
\includegraphics[width=\textwidth]{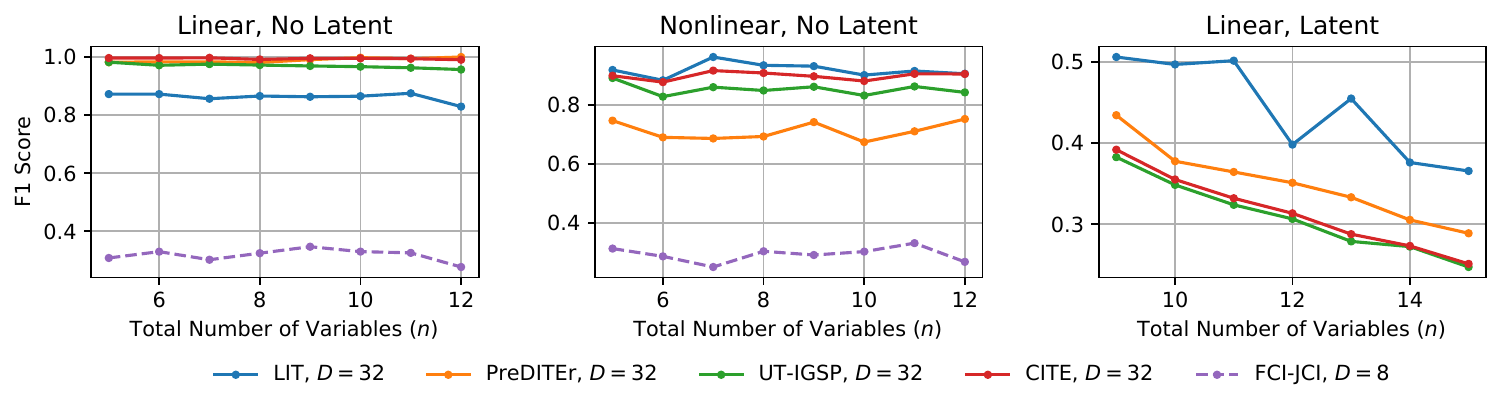}
\caption{Comparison of LIT algorithm with previous work in locating intervention targets.
}
\label{fig:simulation} 
\end{figure}

\section{Experiments}
We evaluated the performance of LIT algorithm on randomly generated models.\footnote{Our code is available at: \url{https://github.com/Yuqin-Yang/LIT}.} Specifically, we considered the following three \new{settings for data generation, and each with number of environments $D=\{8, 16, 32\}$}:
    (1) Linear Gaussian model under causal sufficiency assumption;
    (2) Nonlinear model under causal sufficiency assumption;
    (3) Linear Gaussian model in the presence of latent confounders.
We considered the following approaches in our empirical studies: 1- LIT (our proposed method); 2- PreDITEr algorithm \citep{varici2022intervention}, which allows for latent confounders (that are not in $\Tv$) but assumes the model is linear; 
3- UT-IGSP algorithm \citep{squires2020permutation}, which works for both linear and nonlinear SCMs but under causal sufficiency;
\new{4- CITE algorithm in \citep{varici2021scalable}, which only works for linear SCMs under causal sufficiency;}
5- FCI-JCI123 algorithm in \citep{mooij2020joint}, which allows for both latent confounders as well as nonlinearity in the model.

We repeated each setting for 40 times, and reported the average F1-score in recovering $\Tv_{\Ov}$ for each setting. The results are shown in Figure \ref{fig:simulation}. (Full experimental results can be found in Appendix \ref{app:exp}.) Note that FCI-JCI is executable only under the first two settings with $D=8$ due to huge run times. The results of other algorithms for $D=8$ and $D=16$ are provided in Appendix \ref{app:exp}.
In the first setting, as we expected, PreDITEr and CITE have the best performances as they are designed specifically for linear Gaussian SCMs. UT-IGSP and LIT algorithm have decent performances, while FCI-JCI123 does not perform well. In the second setting, LIT and CITE can both recover the intervention targets with high accuracy. On the contrary, the performance of PreDITEr becomes worse as it is not designed for nonlinear models. Finally, in the third setting, in the presence of latent confounders, the performance of UT-IGSP and CITE becomes much worse because they cannot handle any latent confounders. Meanwhile, LIT outperforms baseline algorithms for various numbers of variables in the system. Note that the best F1-score that can be achieved by any algorithm is strictly less than one as there are observed variables that are theoretically indistinguishable from the true intervention targets.
Lastly, we note that LIT algorithm significantly reduces the number of CI tests performed: LIT algorithm takes at most 80 CI tests while PreDITER requires up to 30000 PDE estimates.
See Appendix \ref{app:exp} for more details.

\section{Comparison with previous work}
\label{sec:related work}
Learning causal structures in multi-environment setting with unknown target interventions has been the focus of several recent work. 
Under the causal sufficiency assumption, \cite{ghassami2017learning} considered linear SCMs in multi-environment setting and showed that the intervention targets can be recovered by checking whether the variances of residuals pertaining to certain linear regressions change across the environments.
\cite{squires2020permutation} proposed an algorithm to recover an interventional Markov equivalence class (I-MEC) from a collection of interventional distributions with unknown intervention targets. The proposed algorithm greedily searches over the space of permutations to minimize a score function. In computing the score function, it is required to perform invariance tests to check whether pairs of distributions are equal which is in general sample inefficient in practice. \cite{brouillard2020differentiable} proposed a causal discovery algorithm that recovers the intervention targets by finding the optimal solution of an appropriately defined score function. In both \citep{squires2020permutation,brouillard2020differentiable},  no interventions are allowed in the observational  environment and therefore the algorithm must know which environment pertains to the observation-only setting. More recently, \citet{perrycausal} proposed a score-based approach based on sparse mechanism shift hypothesis and showed that it can recover the true causal graph up to an I-MEC. However, it has the same drawback of performing invariance tests as in \citep{squires2020permutation}.

In the presence of latent confounders, \cite{jaber2020causal} defined $\Psi$-Markov property which connects the collection of interventional distributions in the multi-environment setting to a pair of a causal graph and set of intervention targets. Moreover, they proposed a sound and complete algorithm to learn the equivalence class of all pairs of graphs and intervention targets that are consistent with the interventional distributions. The proposed algorithm can be used to learn a superset of variables that are intervened between any pair of environments. If there is no latent confounder in the system, this algorithm recovers the true intervention targets.
Similar to \citep{squires2020permutation}, the proposed algorithm requires performing invariance tests to check whether pairs of distributions are equal across environments. Moreover, it needs an exponentially growing number of conditional independence and invariance tests as the number of variables increases.
\cite{mooij2020joint} proposed a causal modeling framework that considers an auxiliary context variable for each environment and applies standard causal discovery algorithms (such as FCI) to learn the causal relationships among context variables and system variables. They provided graphical conditions to read off the intervention targets from the output of their algorithm. 

All the aforementioned work learns the causal structure up to an equivalence class and identifies a candidate intervention set as a byproduct of the output of the causal discovery task. 
Very recently, \cite{varici2021scalable,varici2022intervention} proposed two methods to learn merely intervention targets (rather than recovering the causal structure) in linear Gaussian SCMs with or without latent confounders. Their methods are more scalable than previous work as they do not perform extensive CI tests. Moreover, the identifiability result in recovering intervention targets is exactly the same as \citep{jaber2020causal}.  Note that all the previous work \citep{jaber2020causal,mooij2020joint,varici2022intervention} assume that if latent confounders exist, none of them is in the intervention target set. We relax this assumption and allow latent confounders to be among the intervention targets.
\section{Conclusions}
\label{sec:con}
We addressed the problem of identifying unknown intervention targets in a multi-environment setting. In particular, 
we considered the case when latent confounders are allowed to be in the intervention target set. This may happen when we can only partially observe the system variables. For instance, this is the case in genomics data, where it might be the case that not all affected genes or proteins are measured \citep{verny2017learning}. Existing approaches such as $\psi$-FCI and PreDITEr do not allow for changing latent confounders and lack theoretical guarantees in such scenarios.
Our two-phase algorithm recovers the exogenous noises and matches them with corresponding endogenous variables. Under the causal sufficiency assumption, our algorithm uniquely identifies the intervention targets.
In the presence of latent confounders, we provided a candidate intervention target set which is more informative than previous work. 
Experiment results support the advantages of the proposed algorithm in identifying intervention targets. Our simulations showed existing methods might misinterpret changes in observed variables influenced by these latent confounders, leading to lower precision. In contrast, our method effectively handles latent confounders in intervention targets. As a future work, in the recovery phase, it is an interesting direction to strengthen the identifiability result of non-linear SCM with latent variables which would broaden the applicability of our method to more complex systems.

\subsubsection*{Acknowledgements}
Negar Kiyavash’s research was in part supported by the Swiss National Science Foundation under NCCR Automation, grant agreement 51NF40\_180545 and Swiss SNF project 200021\_204355 /1.

\bibliographystyle{apalike}
\bibliography{ref.bib}

\newpage


\appendix

\section{Further discussion on the recovery phase}\label{app:recovery}
\subsection{Detailed Description about Contrastive Learning Approach and Nonlinear ICA}
Nonlinear ICA refers to an instance of unsupervised learning, where the goal is to learn the independent components/features that generate multi-dimensional observed data. In particular, suppose that $\Xv=(X_1,\cdots,X_n)$, is a $n$-dimensional vector that is generated from $n$ independent components $\Nv=(N_1,\cdots,N_n)$. Let $\gv:\mathbb{R}^n\rightarrow\mathbb{R}^n$ be a smooth and invertible function transforming the latent components (aka sources) to the observed data, i.e., $\Xv=\gv(\Nv)$. Function $\gv$ is called the ``mixing function''. The goal in nonlinear ICA is to recover the inverse function $\gv^{-1}$ and also the latent components in $\Nv$. 

We briefly describe the general approach in recent advances in nonlinear ICA \citep{hyvarinen2019nonlinear,khemakhem2020variational,sorrensondisentanglement} for the case where no additional assumptions are made about the class of mixing functions.  The main idea is to exploit non-stationarity in the data to recover the independent components. In particular, each component $N_i$ depends on some auxiliary variable $U$ and it is independent of other components given $U$, i.e., $\log p(\Nv|U)= \sum_i q_i(N_i|U),$
where $q_i$s are some functions. 
The auxiliary variable $U$ could be an index of a time segment or the index of some environments where we obtained samples from variables in $\Xv$. In this formulation, the distributions of components can change across the environments or time segments. It is often assumed that the distribution of each $N_i$ given $U$ is a member of the exponential family.
\begin{definition}
    A random variable $N_i$ belongs to the  exponential family of order one given a random variable $U$ if its conditional probability distribution function (pdf) can be written as: $ p(N_i|U)=\frac{Q_i(N_i)}{Z_i(U)}\exp\left( \lambda_{i}(U) \tilde{q}_{i}(N_i)\right)$
    where $Q_i$, $Z_i$, $\lambda_{i}$s, and $\tilde{q}_{i}$s are some scalar-valued functions. 
    \label{def:exp family}
\end{definition}
\begin{example}
    For $\tilde{q}_{i}(N_i)=-N_i^2/2$, and $Q_i(N_i)=1$,  the above conditional pdf reduces to Gaussian whose variance is changing across the environments. 
\end{example}
The general approach to  exploit non-stationarity in data is to use contrastive learning to transform the unsupervised learning problem in nonlinear ICA to a supervised learning task. Specifically, a classifier is trained to discriminate samples of a real dataset from their randomized version, i.e., $ \tilde{\Xv}=(\Xv,U)$ versus $\tilde{\Xv'}=(\Xv,U')$,
where $U'$ is drawn randomly from the distribution of $U$, which in practice can be obtained by randomized permutations of the samples of $U$. 

In this approach, a nonlinear regression model is trained with the following form: $r(\Xv,U)= \hv(\Xv)^T\vv(U)+ a(\Xv) + b(U)$
where $\hv(\Xv):\mathbb{R}^n\rightarrow\mathbb{R}^{n}$, $\vv(U):\mathbb{R}\rightarrow\mathbb{R}^{n}$, and $a,b$ are some scalar-valued functions. The model classifies a sample coming from the real data set with probability $1/(1+\exp(-r(\Xv,U)))$.
It has been shown in several work such as in \citep{hyvarinen2019nonlinear,khemakhem2020variational} that if all the components are changed enough across the environments, then the independent components can be recovered from $\hv(\Xv)$ up to some permutation and component-wise nonlinear transformation.

\paragraph{How constrastive learning approach above is applied in our work.}
As we have access to a collection of distributions in $\mathcal{P}$, we will exploit the heterogeneity in data to recover the exogenous noises in $\Nv_{\Tv}$. Let auxiliary variable $U$ denote the index of the environment and assume that the exogenous noises belong to the exponential family in Definition \ref{def:exp family}. Moreover, assume that $\lambda_{i}$'s corresponding to any $N_i\in \Nv_{\Tv}$ are randomly generated across the environments and  $\tilde{q}_{i}(N_i)$s are strictly monotonic functions of $|N_i|$. 
We utilize a contrastive learning approach similar to what we discussed above. We train the nonlinear regression model 
    $r(\Ov,U)= \hv(\Ov)^T\vv(U)+ a(\Ov) + b(U)$
    with universal approximation capability for the supervised learning task of discriminating $(\Ov,U)$ from $(\Ov,U')$ where $\hv(\Ov):\mathbb{R}^{|\Ov|}\rightarrow\mathbb{R}^{|\Ov|}$, $\vv(U):\mathbb{R}\rightarrow\mathbb{R}^{|\Ov|}$, and $a(\Ov)$, $b(U)$ are some scalar-valued functions.

\subsection{Additional Discussion on Assumption 1}
\paragraph{Assumption 1(a).} We provide two examples on when Assumption \ref{assump:tidleg}(a) is satisfied. Note that Assumption \ref{assump:tidleg}(a) requires that $|\Ov|=n=|\Xv|$, which indicates that the model is causally sufficient. 

First, for linear SCMs, the structural equations can be written in vector form as $\Xv=\Bv \Xv+\Nv$ where $\Bv$ is $n\times n$ matrix. Rewrite this equation as $\Xv=(\Iv-\Bv)^{-1}\Nv$. Therefore, the corresponding mixing function is given by $(\Iv-\Bv)^{-1}$ and the above assumption is satisfied if and only if $\Iv-\Bv$ is invertible, which is already satisfied when the causal graph is a DAG. 

Second, for nonlinear SCMs, the invertibility of the mixing function is satisfied when the model is an additive noise model, which can be written as $X_i=f_i(Pa_i)+N_i$. In this case, the inverse function that maps $\Nv$ to $\Ov$ can be constructed according to the following equation: $N_i=X_i - f_i(Pa_i)$. Note that the invertibility of the mixing function does not depend on $f_i$. \new{This result can be generalized into the following remark:
\begin{remark}
    For a nonlinear SCM $X_i=f_i(Pa_i, N_i)$, Assumption \ref{assump:tidleg}(a) is satisfied if the model is acyclic, $f_i$ is continuous, and the partial derivative $\partial f_i / \partial N_i$ is strictly negative or positive for all $X_i\in \Xv$ and any values of $Pa_i$.
\end{remark}
We note that the condition in Remark 3 can be satisfied when $f_i$ in the data generating model is designed as a multi-layer perceptron with ReLU activation function, where all model weights are strictly positive. 
}

\paragraph{Assumption 1(b).} For Assumption \ref{assump:tidleg}(b), we note that if the SCM is a linear non-Gaussian model, i.e., linear SCM with non-Gaussian exogenous noises, and latent confounders are present, then Assumption \ref{assump:tidleg}(b) implies that: (1) All intervention targets must be observed variables, i.e., $\Tv\subseteq \Ov$; and (2) Latent variables cannot have children in $\Tv$. In other words, $\Tv$ can only include observed variables that do not have latent parents. Please see Appendix \ref{app:proof_remark3} for the proof. 
\new{However, we observed experimentally that if the SCM is linear Gaussian within each environment, then we can recover the noises in $\Nv_\Tv$ while allowing latent confounders to be in $\Tv$ using linear ICA methods.}

\section{Further discussion on the matching phase}
\subsection{Markov Property in the Augmented Graph $G_{\Tv}$}
\label{proof: Markov property}
For the given SCM $\mathcal{M}$, we construct an augmented version $\mathcal{\tilde{M}}_{\Tv}$ as follows. We add $\tilde{\Nv}_{\Tv}$ to the set of endogenous variables in $\mathcal{M}$ and remove $\Nv_{\Tv}$ from the set of exogenous noises. 
Moreover, for any $X_i\in \Tv$, we change its structural assignment as follows: $X_{i}:=\tilde{f}_{i}(\PA_{i},\tilde{N}_{i})=f_{i}(\PA_{i},g_{i}(\tilde{N}_{i}))$ where $\tilde{N}_i$ is the corresponding recovered noise in $\tilde{\Nv}_{\Tv}$, $\tilde{f}_{i}$ is the new causal mechanism of $X_{i}$ relating it to its new set of parents $\PA_i\cup \{\tilde{N}_i\}$ and $g_{i}$ is an invertible function such that $N_{i}=g_{i}(\tilde{N}_{i})$. For the variables that are not in $\Tv$, we keep their corresponding structural assignments unchanged. Please note that with this construction, the joint distribution over $\Xv$ entailed by SCM $\tilde{\mathcal{M}}_{\Tv}$ is exactly the same as the one entailed by original SCM $\mathcal{M}$. With the exact same argument in the proof of Theorem 1.4.1 in \citep{pearl2009causality}, it can be shown that the distribution $p(\Xv,\tilde{\Nv}_{\Tv})$ entailed by SCM $\tilde{\mathcal{M}}_{\Tv}$ satisfies the local Markov property as the value of each observed variable is uniquely determined given the values of its parts and the corresponding exogenous noise. Moreover, in causal DAGs, the local Markov property implies the Global Markov property \citep{geiger1990logic}. Hence, the joint distribution $p(\Xv,\tilde{\Nv}_{\Tv})$ satisfies Markov property with respect to its corresponding causal graph, $G_{\Tv}$.

In the presence of latent confounders, we modify $\mathcal{\tilde{M}}_{\Tv}$ as follows. For any $X_l\in \Tv_{\Lv}$, we consider it as an observed variable and replace its structural assignment with $X_l:=\tilde{N}_{l'}$ where $\tilde{N}_{l'}$ is the corresponding recovered noise. Moreover, for any other assignment that $X_l$ appears, we replace it with $g_l(X_l)$. Similar to the causally sufficient case, the distribution $p(\Xv,\tilde{\Nv}_{\Tv})$ entailed by SCM $\tilde{\mathcal{M}}_{\Tv}$ satisfies the local Markov property since the value of each observed variable is still uniquely determined given the values of its parents and the corresponding exogenous noise. 

\subsection{Example Explaining the Conditions under the Causal Sufficiency Assumption}\label{app:example}
The following example illustrates how the the conditions in Proposition \ref{th:main_condition} can be used to recover $\Tv$.
\begin{wrapfigure}{r}{0.35\textwidth}
    \centering
    \includegraphics[width=0.35\textwidth]{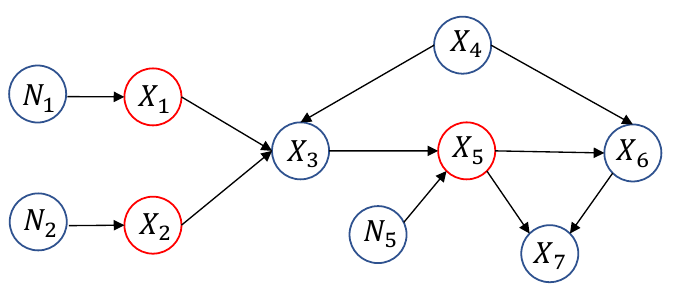}
    \caption{An example of SCM:  Intervention targets $\Tv=\{X_1,X_2,$ $X_5\}$ are shown by red circles.}
    \label{fig:example_1}
\end{wrapfigure}
\begin{example}
Figure \ref{fig:example_1} depicts the augmented graph of an SCM in which $\Tv=\{X_1,X_2,X_5\}$ (indicated by red circles). In the recovery phase, we recover three noises $\tilde{N}_1,\tilde{N}_2,\tilde{N}_5$, which are invertible translations of $N_1$, $N_2$, $N_5$, respectively. Note that we do not know the correspondence of the noises to the variables as there are permutation indeterminacy.
The indicator sets for all the variables are: $\Iv_1=\{\tilde{N}_1\}$, $\Iv_2=\{\tilde{N}_2\}$, $\Iv_3=\{\tilde{N}_1,\tilde{N}_2\}$, $\Iv_4=\emptyset$, $\Iv_5=\Iv_6=\Iv_7=\{\tilde{N}_1,\tilde{N}_2,\tilde{N}_5\}$.
For $X_1$ and $X_2$, the condition in \ref{proposition-2} is satisfied. Thus, they are in $\Tv$. As for $X_3$ and $X_4$, the condition in \ref{proposition-1} holds and therefore they are not in $\Tv$. For the variables in $\{X_5,X_6,X_7\}$, the condition in \ref{proposition-3} holds and the only variable satisfying the condition in \eqref{eq:third_cond} is $X_5$ as $\tilde{N}_5$ is independent of $X_6$ and $X_7$ given $\Sv_5\cup\{X_5\}$ where $\Sv_5=\{X_1,X_2,X_3,X_4\}$.
\end{example}

\section{Efficient algorithm based on mixing matrix}
\label{app:efficient alg}
We show that the problem of locating the intervention targets in the causally sufficient systems can be reduced to the problem of causal discovery on linear SCMs with deterministic relations \citep{yang2022causal, yang2022causal1}. This allows us to design a more efficient algorithm based on existing causal discovery algorithms for linear SCMs. 
In the following, we first show that there is a connection between the indicator set $\mathcal{I}$ and the mixing matrix of a linear SCM. Next, we present an alternative algorithm of LIT under causal sufficiency based on existing algorithms, and analyze its computational complexity. Lastly, we extend the algorithm to the case with latent confounders.

To show these connection between the two problems, we define a mapping $\phi$ from the set of SCMs with soft interventions and the set of linear SCMs with deterministic relations, where for each SCM $\setM$ with intervention target set $\Tv$, $\phi(\setM, \Tv)$ is constructed as follows. Each variable $X$ in $\setM$ is mapped to a non-deterministic variable, denoted by $\phi(X)$, if $X\in\Tv$, and is mapped to a deterministic variable if it is not in $\Tv$. Further, for each pair of variables $(X_i, X_j)$ in $\setM$, $\phi(X_i)$ is a parent of $\phi(X_j)$ in $\phi(\setM)$ with coefficient 1 if and only if $X_i$ is a parent of $X_j$ in $\setM$. 
The following proposition states that the indicator set of $\setM$ is the same as the support of the mixing matrix of $\phi(\setM, \Tv)$.
\begin{proposition}
\label{prop:mapping}
For a given SCM $\setM$ with intervention target set $\Tv$, we denote 
the $(i,j)$-th entry of the mixing matrix of $\phi(\setM, \Tv)$ as $w_{ij}$. Then for all $i\in [n]$ and $j\in [|\Tv|]$, $w_{ij}\neq 0$ if and only if $\tilde{N}_j\in \Iv_i$.
\end{proposition}

\begin{remark}
Proposition \ref{prop:mapping} implies that if the underlying model is linear Gaussian, then under Assumption \ref{assump:faithfulness}, the indicator set can be obtained from the recovered mixing matrix in linear ICA.
\end{remark}

\begin{example}
Consider SCM $\setM$ over a collider structure: $X_1\rightarrow X_3 \leftarrow X_2$, where $X_1$ and $X_2$ are intervention targets with corresponding exogenous noises $\tilde{N_1}$ and $\tilde{N_2}$, respectively. The indicator sets are: $\Iv_1=\{\tilde{N}_{1}\}$, $\Iv_2=\{\tilde{N}_2\}$, and $\Iv_3=\{\tilde{N}_1,\tilde{N}_2\}$. Under the mapping $\phi$, $\setM$ is mapped to the linear SCM $\setM'$ over the same collider structure, where $\phi(X_3)$ is a deterministic variable, and $\phi(X_1), \phi(X_2)$ are non-deterministic. The mixing matrix corresponding to $\setM'$ is $\Wv=[1\; 0;\; 0\; 1;\; 1\; 1]$, where for all $i=1,2,3$, $\Iv_i$ represents the support of the $i$-th row of $\Wv$.
\end{example}


Proposition \ref{prop:mapping} implies that any recovery algorithm for linear SCMs with deterministic relations, that are based on the mixing matrix, can be applied to our problem. In particular, \cite{yang2022causal} considered the problem of causal discovery on linear SCMs with measurement error, which can be considered as a special case of deterministic relations. They proposed AOG recovery algorithm, where the input is the mixing matrix, and the output is a partition of the variables into distinct sets such that variables in the same set has the same row support. Note that the conditions in \ref{proposition-1} and \ref{proposition-2} in Proposition \ref{th:main_condition} is purely based on indicator set, and the condition \ref{proposition-3} is based on partitioning variables based on the indicator set. 

We devise a more efficient version of the LIT algorithm in Algorithm \ref{alg:aog} which utilizes AOG recovery algorithm. Specifically, instead of iterating over all variables, Algorithm \ref{alg:aog} iterates over the recovered noises, which corresponds to the non-empty residual sets $\setN_i$ in Proposition \ref{th:main_condition}. (Recall that under causal sufficiency assumption, $|\setN_i|=1$ if it is not empty.)
During each iteration, the algorithm finds a row $X_i$ with only one non-zero entry and has the smallest corresponding value in $\mathbf{n}$. Note that this entry corresponds to the noise in $\setN_i$, and for any variable $X_j$ that has the same support on the submatrix as $X_i$, if the number in $\mathbf{n}$ corresponding to $X_j$ is larger than the number corresponding to $X_i$, \saber{then $\Iv_i$ is unique and $X_i$ satisfies condition \ref{proposition-2}. Moreover, $\Iv_i$ is a strict subset of $\Iv_j$ and thus $X_i\in \Sv_j$. Please note that $X_j$ satisfies the condition in \ref{proposition-1} since the union of the indicator sets for $X_i$ and a subset of selected variables in the previous loops is equal to $\Iv_j$. Therefore, $X_j$ should be excluded from $\Tv$.
}
Otherwise, if the number in $\mathbf{n}$ corresponding to $X_j$ is equal to the number corresponding to $X_i$, then $\Iv_i=\Iv_j$, and we need to use \eqref{eq:third_cond} to find the variable in $\Kv$. 

To show that Algorithm \ref{alg:aog} is a more efficient version of Algorithm \ref{alg1}, note that the first step of Algorithm \ref{alg1} is to compute $\Sv_i$ for all observed variables $X_i$. Since $\Sv_i$ involves pairwise comparison of the indicator sets, the time complexity of this step is $\Theta(n^2|T|)$, where $n$ is the number of observed variables, and $|\Tv|$ is the number of recovered exogenous noises. 
On the contrary, Algorithm \ref{alg:aog} does not need to calculate all $\Sv_i$. It is only needed when we need to check \eqref{eq:third_cond}. Since all variables in $\setX_{i}$ have the same indicator set, finding $\Sv_i$ takes at most $\Theta(n|\Tv|^2)$.  
It can be shown that the time complexity for Algorithm \ref{alg1} is $\Theta(n^2|\Tv|)$, as the conditions in \ref{proposition-1} and \ref{proposition-2} are hard to check and each takes $\Theta(n^2|\Tv|)$ time. On the contrary, the time complexity for Algorithm \ref{alg:aog} is  $\Theta(n|\Tv|^2)$.

\begin{algorithm}[t]
\SetAlgoVlined
\LinesNumbered
\DontPrintSemicolon
Obtain $\TNT$ and $\mathcal{I}$ from data. Rewrite $\mathcal{I}$ into the matrix form, where each row represents an indicator set. Denote it as $\Iv$.\;
Calculate the number of non-zero entries in each row of $\bI$. Denote the vector of these numbers as $\mathbf{n}$, where each entry in $\mathbf{n}$ corresponds to a row in $\bI$.\; 
Initialize $\tilde{\bI}=\bI$, $\Kv\gets \emptyset$.\;
Remove all rows in $\tilde{\bI}$ with no non-zero entry.\;
\While{$\tilde{\bI}$ is not empty}{
Find a row in $\tilde{\bI}$ that contains only one non-zero entry, and has the smallest corresponding value in $\mathbf{n}$. If there are multiple such rows, randomly select one. Denote the selected row as $\bw$, and its corresponding value in $\mathbf{n}$ as $n_0$. \; 
Consider the rows in $\tilde{\bI}$ with the same support (non-zero entry) as $\bw$, including $\bw$ itself. Denote the set of variables that correspond to these rows as $\mathcal{Z}_I$.\;
Denote the set of variables in $\mathcal{Z}_I$ with the same corresponding value as $n_0$ in $\mathbf{n}$ as $\mathcal{Z}_J$.\; 
\eIf{$\mathcal{Z}_J$ has only one element}{
Add the variable that corresponds to $\bw$ to $\Kv$.\;
}
{
Find the only variable $X_{i_k}$ in $\mathcal{Z}_J$ that satisfies \eqref{eq:third_cond}; $\Kv\gets \Kv\cup \{X_{i_k}\}$.\;
}
Remove from $\tilde{\bI}$ the rows corresponding to the variables in $\mathcal{Z}_I$, and the column containing the corresponding non-zero entries in these rows.\;
}
\Return{$\Kv$}
\caption{Locating interventions target (LIT) algorithm under causal sufficiency.}
\label{alg:aog} 
\end{algorithm}

Lastly, we can extend Algorithm \ref{alg1} to the case in the presence of latent confounders. The new algorithm is shown in Algorithm \ref{alg:aog-latent}. 
In particular, in order to check \ref{proposition-1-L}, we count the number of non-zero entries in each column, and check \ref{proposition-1-L} in lines 8-11, which is also a condition that is purely based on the indicator sets.

\begin{algorithm}[t]
\SetAlgoVlined
\LinesNumbered
\DontPrintSemicolon
Obtain $\TNT$ and $\mathcal{I}$ from data and rewrite $\mathcal{I}$ into the matrix form $\Iv$.\;
Calculate the number of non-zero entries in each row and each column of $\bI$. Denote the vector of these numbers as $\mathbf{n}$ and $\mathbf{m}$, where each entry in $\mathbf{n}$ corresponds to a row in $\bI$, and each entry in $\mathbf{m}$ corresponds to a column in $\bI$.\; 
Initialize $\tilde{\bI}=\bI$, $\Kv \gets \emptyset$.\;
Remove all rows in $\tilde{\bI}$ with all entry being zero.\;
\While{$\tilde{\bI}$ is not empty}{
Find a row in $\tilde{\bI}$ that contains the fewest number of non-zero entry, and has the smallest corresponding value in $\mathbf{n}$. If there are multiple such rows, randomly select one. Denote the selected row as $\bw$, the variable corresponding to this row as $X_w$, and the corresponding value in $\mathbf{n}$ as $n_0$.
\; 
Consider the rows in $\tilde{\bI}$ with the same support (non-zero entry) as $\bw$, including $\bw$ itself. Denote the set of variables that correspond to these rows as $\mathcal{Z}_I$, and the the noise terms corresponding to the support of $\bw$ as $\setN_I$.\;
Denote the set of variables in $\mathcal{Z}_I$ with corresponding value $n_0$ in $\mathbf{n}$ as $\mathcal{Z}_J$, and the set of noise terms in $\mathcal{N}_I$ with the smallest corresponding value in $\bmm$ as $\mathcal{N}_J$. \; 
Randomly select one noise term $N_m$ in $\setN_J$. Consider the submatrix of $\bI$ where the rows correspond to the (column) support of $N_m$, and the columns correspond to the (row) support of $X_w$. \;
\eIf{This submatrix includes any zero entry}{\textbf{pass}}{
\If{$|\mathcal{Z}_J|> 1$}{
Find the set of variables in $\setZ_J$ that satisfy \eqref{eq:third_cond_latent}. Denote it as $\setZ_K$.\;
$\setZ_J \gets \setZ_J\setminus\setZ_K$.\;
}
$\Kv\gets \Kv\cup \setZ_J$.\; 
}
Remove from $\tilde{\bI}$ the rows corresponding to the variables in $\mathcal{Z}_I$, and the columns corresponding to the noise terms in $\setN_I$.\;
}
\Return{$\Kv$}
\caption{LIT algorithm in the presence of latent confounders.}
\label{alg:aog-latent} 
\end{algorithm}

\section{Proofs}
\subsection{Proof of Proposition \ref{prop:recover_sources}}
We first prove the statement of proposition for the case of $\Lv=\emptyset$. With infinite samples and a model with universal approximation capability, after training, the regression model will equal the difference of the log-densities in the two classes:
\begin{equation}
\begin{split}
    r(\Xv,U) &= \log p(\Nv,U) + \log |\Jv \gv (\Xv)| -\log p(\Nv) - \log p(U) - \log |\Jv \gv (\Xv)|\\
    &= \log p(\Nv|U) - \log p(\Nv)\\
    &= \sum_i \log Q_i(N_i) -\log Z_i(U)+ \tilde{q}_{i}(N_i) \lambda_{i}(U) - \log p(\Nv).
\end{split}
\end{equation}
If we consider the form of $r(\Xv,U)= \hv(\Xv)^T\vv(U)+ a(\Xv) + b(U)$ for $r(\Xv,U)$, we can set the functions $\hv(\Xv)$, $\vv(U)$, $a(\Xv)$, and $b(U)$ such that it is equal to the right hand side of above equation. In particular, we can have the following equality 
\begin{equation}
    \sum_{i} h_{i}(\Xv) v_{i}(U) + a(\Xv) + b(U) = \sum_i \log Q_i(N_i) -\log Z_i(U)+ \sum_{i} \tilde{q}_{i}(N_i) \lambda_{i}(U) - \log p(\Nv),
\end{equation}
with the following possible solution:
\begin{equation}
\begin{split}
    h_{i}(\Xv)=\tilde{q}_{i}(N_i),
    v_{i}(U)=\lambda_{i}(U),a(\Xv)=\sum_i \log Q_i(N_i)-\log p(\Nv), b(U)= -\sum_i \log Z_i(U).
\end{split}
\end{equation}
The random variable $U$ is equal to $d$, if the sample is drawn from the environment $E_d$ where $1\leq d \leq D$. Let $\Pv$ be $D\times n$ matrix where $d$-th row is equal to $[\lambda_{1}(d),\cdots,\lambda_{n}(d)]$. We  collect $\tilde{q}_{i}$s in the vector $\tilde{\qv}(\Nv)=[\tilde{q
}_{1}(N_1),\cdots,\tilde{q}_{n}(N_n)]^T$. We also define the matrix $\Vv$ where $d$-th row is equal to $[v_{1}(d),\cdots,v_{n}(d)]$. Finally, we collect $-\sum_i\log Z_i(U)-b(U)$ for different values of $U$ in a vector $\Zv$. Based on these definitions, we have:
\begin{equation}
    \Vv \hv(\Xv)= \Pv \tilde{\qv}(\Nv) + \Zv +\mathbf{1}\left(\sum_i \log Q_i(N_i)-\log p(\Nv)-a(\Xv)\right),
\end{equation}
where $\mathbf{1}$ is a $D\times 1$ vector of all ones. If from both sides of the above equation, we subtract the first row from the others,
\begin{equation}
    \Vv' \hv(\Xv) = \Pv' \tilde{\qv}(\Nv)+\Zv',
\end{equation}
where $\Vv',\Pv'$, and $\Zv'$ denote the resulting matrices after subtraction corresponding to $\Vv,\Pv,$ and $\Zv$, respectively.

The columns corresponding to exogenous noises that are not changing across environments are zeros in $\Pv'$. Thus, we can remove these columns from $\Pv'$ and also the corresponding entries in $\tilde{\qv}(\Nv)$. We denote the resulting matrix and vector by $\Pv''$ and $\tilde{\qv}(\Nv_{\Tv})$, respectively. Hence, we can rewrite the above equation as follows:
\begin{equation}
    \Vv' \hv(\Xv) = \Pv'' \tilde{\qv}(\Nv_{\Tv})+\Zv'.
\end{equation}
As $\lambda_{i}(U)$s are generated randomly across the environments and $D\geq |\Tv|+1$, $\Pv''$ is full column rank with measure one. Therefore, we have:
\begin{equation}
    \tilde{\qv}(\Nv_{\Tv})= (\Pv'')^{\dagger} \Vv' \hv(\Xv) - \Zv'',
\end{equation}
where $(\Pv'')^{\dagger}$ is pseudo-inverse of matrix $\Pv''$ and $\Zv'' = (\Pv'')^{\dagger} \Zv'$. Since we know that the entries of $\tilde{\qv}(\Nv_{\Tv})$ are linearly independent, $(\Pv'')^{\dagger} \Vv'$ is full row rank. Moreover, $q_i(N_i)$s are non-Gaussian as they are bounded from above to ensure integrability. Thus, we can recover $\tilde{\qv}(\Nv_{\Tv})$ from $\hv(\Xv)$ by solving an under-complete linear ICA problem.

Now, let us assume that there are some latent variables in the system, i.e., $\Lv\neq \emptyset$. As we know that there exists an invertible function $\tilde{\gv}$ such that $\tilde{\gv}(\Ov)=(\Nv_{\Tv}, \Vv)$, we have:
\begin{equation}
    \begin{split}
    r(\Ov,U) &= \log p(\Ov,U) -\log p(\Ov) - \log p(U) \\
    &= \log p(\Ov|U) - \log p(\Ov)\\
    &= \log p(\Nv_{\Tv},\Vv|U) + \log |\Jv \tilde{\gv} (\Ov)| - \log p(\Nv_{\Tv},\Vv) - \log |\Jv \tilde{\gv} (\Ov)|\\
    &=\log p(\Nv_{\Tv}|U) - \log p(\Nv_{\Tv}), 
\end{split}
\end{equation}
where the third equality is due to the existence of invertible function $\tilde{\gv}$ and the last equality is according to the assumptions that $\Nv_{\Tv}\indep \Vv|U$, $\Vv\indep U$. Please note that these two assumptions imply that $\Nv_{\Tv}\indep \Vv$. Similar to the causally sufficient case, based on the form of $r(\Ov,U)$, we can write the following equation:
\begin{equation}
    \sum_{j} h_{j}(\Ov) v_{j}(U) + a(\Ov) + b(U) = \sum_{i:X_i\in \Tv} \log Q_i(N_i) -\log Z_i(U)+ \tilde{q}_{i}(N_i) \lambda_{i}(U) - \log p(\Nv_{\Tv}).
\end{equation}

where $h(\Ov):\mathbb{R}^{|\Ov|}\rightarrow \mathbb{R}^{|\Ov|}$. Let $\Mv$ be $D\times |\Tv|$ matrix where $d$-th row is equal to $[\lambda_{1}(d),\cdots,\lambda_{|\Tv|}(d)]$. We  collect $\tilde{q}_{i}$s in the vector $\tilde{\qv}(\Nv_{\Tv})$. We also define the matrix $\Wv$ where $d$-th row is equal to $[v_{1}(d),\cdots,v_{|\Ov|}(d)]$. Finally, we collect $-\sum_{i:X_i\in \Tv}\log Z_i(U)-b(U)$ for different values of $U$ in a vector $\Zv$. Based on these definitions, we have:
\begin{equation}
    \Wv \hv(\Ov)= \Mv \tilde{\qv}(\Nv_{\Tv}) + \Zv +\mathbf{1}\left(\sum_{i:X_i\in \Tv} \log Q_i(N_i)-\log p(\Nv_{\Tv})-a(\Ov)\right),
\end{equation}
where $\mathbf{1}$ is a $D\times 1$ vector of all ones. 

Now, if from both sides of the above equation, we subtract the first row from the others, we have
\begin{equation}
    \Wv' \hv(\Ov) = \Mv' \tilde{\qv}(\Nv_{\Tv})+\Zv',
\end{equation}
where $\Vv',\Lv'$, and $\Zv'$ denote the resulting matrices after subtraction corresponding to $\Wv,\Mv,$ and $\Zv$, respectively.

As $\lambda_{i}(U)$s are generated randomly across the environments and $D\geq |\Tv|+1$, $\Mv'$ is full column rank with measure one. Therefore, we have:
\begin{equation}
    \tilde{\qv}(\Nv_{\Tv})= (\Mv')^{\dagger} \Wv' \hv(\Ov) - \Zv'',
\end{equation}
where $(\Mv')^{\dagger}$ is pseudo-inverse of matrix $\Mv'$ and $\Zv'' = (\Mv')^{\dagger} \Zv'$. Since we know that the entries of $\tilde{\qv}(\Nv_{\Tv})$ are linearly independent, $(\Mv'')^{\dagger} \Wv'$ is full row rank. Moreover, $q_i(N_i)$s are non-Gaussian as they are bounded from above to ensure integrability. Thus, we can recover $\tilde{\qv}(\Nv_{\Tv})$ from $\hv(\Ov)$ by solving an under-complete linear ICA problem.

So far, we showed that the recovery phase can be performed up to some component-wise nonlinear transformation (not necessarily an invertible one). However, similar to Corollary 2 in \citep{hyvarinen2016unsupervised}, it can be shown that from $\tilde{\qv}(\Nv_{\Tv})$ and the observed vector $\Ov$, the exogenous noises in $\Nv_{\Tv}$ can be recovered up to some  strictly monotonic transformation if each function $\tilde{q}_{i}(N_i)$ is a strictly monotonic function of $|N_i|$.

\subsection{Regarding Assumption \ref{assump:tidleg}(b) in Linear non-Gaussian Models}
\label{app:proof_remark3}
In the following we prove that, in the linear non-Gaussian model (i.e., linear SCM with non-Gaussian exogenous noises) with the causal graph $G$, if $\Lv\neq\emptyset$, then the conditions in Proposition \ref{prop:recover_sources} imply that: (i) $\Tv\subseteq \Ov$; (ii) For each latent variable $H_i\in \Lv$, $Ch_G(H_i)\cap \Tv = \emptyset$, i.e., $H_i$ cannot have children in $\Tv$ where $Ch_G(H_i)$ is the children of $H_i$. 

The linear SCM has the following matrix form:
\begin{equation}
\Lv = \Nv_\Lv;\quad \Ov = \Av \Ov + \Av' \Lv + \Nv_\Ov, 
\label{eq:linearscm}
\end{equation}
where $\Nv_\Lv$ and $\Nv_\Ov$ represent the vector of exogenous noises associated with latent and observed variables, respectively. We also denote the exogenous noises whose distributions are not changing across the environments by $\Nv_{\Tv^c}$. $\Av$ represents the direct causal relationships among observed variables and $\Av'$ represents the direct causal relations from latent to observed variables. Note that if we permute the variables such that $\Xv=[\Lv,\Ov]$, then the adjacency matrix $\Bv$ after the same (row and column) permutation is $[\mathbf{0}, \mathbf{0}; \Av', \Av]$. Under the acyclicity assumption, $\Av$ can be permuted into a strictly lower triangular matrix. Following  \eqref{eq:linearscm}, $\Ov$ can be written as a linear combination of the noise terms in $(\Nv_\Lv, \Nv_\Ov)$:
\begin{equation}
\Ov = [\Dv\quad \Cv] 
\begin{bmatrix}
\Nv_\Lv \\
\Nv_\Ov
\end{bmatrix},
\label{eq:linearscm_mixing_matrix}
\end{equation}
where $\Dv=(\Iv - \Av)^{-1}\Av'$ and $\Cv=(\Iv - \Av)^{-1}$. Denote $\Wv = [\Dv\quad \Cv]$, which represents the total causal effects (i.e., sum of product of path coefficients) among variables \citep{spirtes2000causation}. If all exogenous noises are non-Gaussian and no two columns of $\Wv$ are linearly dependent of each other, then $\Wv$ can be recovered up to permutation and scaling of the columns using overcomplete ICA. Given that some variables in $\Xv$ belong to $\Tv$, we can rewrite \eqref{eq:linearscm_mixing_matrix} as follows:
\begin{equation}
\Ov = [\Wv_{\Tv}\quad \Wv_{\Tv^c}]
\begin{bmatrix}
\Nv_{\Tv}\\
\Nv_{\Tv^c}
\end{bmatrix},
\label{eq:linearscm_mixing_matrix_t}
\end{equation}
where $\Wv_{\Tv}$ and $\Wv_{\Tv^c}$ represent the submatrix of $\Wv$ that correspond to the exogenous noises in $\Nv_{\Tv}$ and $\Nv_{\Tv^c}$, respectively.

If the conditions in Proposition \ref{prop:recover_sources} hold, then there exists an invertible matrix 
$\Gv$, such that
\begin{equation}
\Gv \Ov =
[\Gv\Wv_{\Tv}\quad \Gv\Wv_{\Tv^c}]
\begin{bmatrix}
\Nv_{\Tv}\\
\Nv_{\Tv^c}
\end{bmatrix}
=
\begin{bmatrix}
\Nv_{\Tv} \\
\Vv
\end{bmatrix}.
\label{eq:linearscm_mixing_matrix_g}
\end{equation}
Partition $\Gv$ into $[\Gv_1; \Gv_2]$, where $\Gv_1\in \mathbb{R}^{|\Tv|\times n}$ represent the first $|\Tv|$ rows of $\Gv$, and $\Gv_2\in \mathbb{R}^{(n-|\Tv|)\times n}$ represent the remaining rows. Therefore, according to  \eqref{eq:linearscm_mixing_matrix_g}, we have
\begin{equation}
\Gv_1\Wv_{\Tv} \Nv_{\Tv} + \Gv_1\Wv_{\Tv^c} \Nv_{\Tv^c}
=
\Nv_{\Tv}.
\label{eq:linearscm_mixing_matrix_g1}
\end{equation}
We first show that if all noises in $\Nv_{\Tv}$ and $\Nv_{\Tv^c}$ are mutually independent and non-Gaussian, then \eqref{eq:linearscm_mixing_matrix_g1} implies that $\Gv_1\Wv_{\Tv}=\Iv$, and $\Gv_1\Wv_{\Tv^c}=\mathbf{0}$. This is because for each noise $N_i\in \Nv_{\Tv}$, $i\in [|\Tv|]$, according to  \eqref{eq:linearscm_mixing_matrix_g1}, $N_i$ can be written as a linear combination of exogenous noises in $\Nv_{\Tv}\cup\Nv_{\Tv^c}$. Since all noises are mutually independent and non-Gaussian, according to Darmois-Skitovitch theorem \citep{Darmois1953AnalyseGD, skitovitch1953property}, the coefficient of any $N_j$, $j\neq i$ on $N_i$ must be zero. Otherwise $N_j$ and $N_i$ are independent, but 

For each exogenous noise $N\in \Nv_{\Tv}\cup\Nv_{\Tv^c}$, denote its corresponding column vector in $\Wv$ in  \eqref{eq:linearscm_mixing_matrix_t} as $\wv_{N}$. Then $\Gv_1\Wv_{\Tv}=\Iv$, $\Gv_1\Wv_{\Tv^c}=\mathbf{0}$ is equivalent to: For each exogenous noise $N$ and its corresponding column vector $\wv_{N}$, $\Gv_1 \wv_N=\mathbf{0}$ if $N\in \Nv_{\Tv^c}$, and $\Gv_1 \wv_N=\ev_N$ if $N\in \Nv_{\Tv}$, where $\ev_N$ is the basis (one-hot) vector where the entry corresponding to $N$ is one and the rest are zero. Further, for each latent variable $H_i\in \Lv$, we have
\begin{equation*}
    \wv_{N_{H_i}} = \sum_{j:X_j\in Ch_G(H)} a'_{ji} \wv_{N_{X_j}}, 
\end{equation*}
where $a'_{ji}$ represent the $(j,i)$-th entry of matrix $\Av'$ in \eqref{eq:linearscm}.
This is because for any observed variable $X\in \Ov$, the total causal effect of $H_i$ on $X$ can be written as summation of the total causal effect from each child of $H_i$ to $X$ multiplied by the direct causal effect from $H_i$ to this child (i.e., $a'_{ji}$). Therefore, we have
\begin{equation}
\Gv_1\wv_{N_{H_i}} = \sum_{j:X_j\in Ch_G(H_i)} a'_{ji} \Gv_1\wv_{N_{X_j}}. 
\label{eq:non-gaussian-final}
\end{equation}
Since $\Gv_1 \wv_N$ corresponds to either zero vector or basis vector, and different noises in $\Nv_\Tv$ correspond to different basis vectors, \eqref{eq:non-gaussian-final} implies that $\Gv_1\wv_{N_{H_i}}=\mathbf{0}$, and $\Gv_1\wv_{N_{X_j}}=\mathbf{0}$ for all latent variable $H_i\in \Lv$, and all $X_j\in Ch_G(H_i)$. This means that $\Tv$ only includes observed variables that do not have latent parents.

\subsection{Proof of Proposition \ref{th:main_condition}}
\label{app:proof of prop 3}
Consider any recovered noise $\tilde{N}_i\in \tilde{\mathbf{N}}_{\Tv}$. Without loss of generality, suppose that $\tilde{N}_i$ corresponds to $X_i$. We first show that $\tilde{N}_i$ only depends on the descendants of $X_i$. First, for any node $X_j\in Des(X_i)$, the path $N_i \rightarrow X_i \rightarrow \cdots \rightarrow X_j$ in the augmented graph $G_{\Tv}$ is not blocked without conditioning on any other variable and hence $\tilde{N}_i \not\!\perp\!\!\!\perp X_j$. Moreover, for any $X_j$ which is a non-descendent of $X_i$, there is always a collider on any path between $X_j$ and $N_i$ and thus it is blocked. Hence, we have: $\tilde{N}_i \!\perp\!\!\!\perp X_j$.

\begin{remark}
Based on what we proved above, we can imply that $\Iv_i$ contains only the noises in $\tilde{\Nv}_{\Tv}$ whose corresponding variables are ancestors of $X_i$ in $\GT$.
\label{remark:1}
\end{remark}




\begin{remark}
For any two variables $X_i,X_j$, if we have $\Iv_j\subsetneq \Iv_i$ , then $X_j$ cannot be a descendent of $X_i$. Since if $X_j$ is a descendent of $X_i$, then based on Remark \ref{remark:1}, we can conclude that $\Iv_i\subseteq \Iv_j$ which violates our assumption. 
\label{remark:2}
\end{remark}

Now, we prove the three statements in the proposition based on the above two remarks:

\ref{proposition-1} By contradiction, suppose that $X_i$ is in $\Tv$. 
If for a variable $X_j$, we have $\Iv_j\subsetneq \Iv_i$, then based on Remark \ref{remark:2}, it cannot be a descendent of $X_i$. Now if the condition in \ref{proposition-1}
satisfies, then there exists a set such as $\Iv_j$, where $j\in \Sv_i$, such that $i\in \Iv_j$. But according to Remark \ref{remark:1}, this means that $X_j$ is a descendant of $X_i$ which is a contradiction.

\ref{proposition-2} As the condition in \ref{proposition-1} is not satisfied, there exists $\tilde{N}_{k'}\in \setN_i=\Iv_i\backslash\cup_{j:X_j\in \Sv_i} \Iv_j$. Suppose that $\tilde{N}_{k'}$ corresponds to $X_k$. Based on Remark \ref{remark:1}, $X_k$ should be an ancestor of $X_i$. Moreover, we have $\Iv_k \subseteq \Iv_i$. As $\Iv_i$ is unique, then $\Iv_k\subsetneq \Iv_i$ and $X_k\in \Sv_i$. But this is in contradiction with the assumption $k'\in \Iv_i\backslash\cup_{j:X_j\in \Sv_i} \Iv_j$ and the proof is complete.

\ref{proposition-3} As the indicator sets do not satisfy the condition in \ref{proposition-1}, $\setN_i=\Iv_{i_1}\backslash \cup_{j\in \Sv_{i_1}} \Iv_j$ is not empty. Suppose that $\tilde{N}_{k'}\in \setN_i$ and $\tilde{N}_{k'}$ corresponds to $X_k$. We know that $X_k$ should be in the set $\setX_i=\{X_{i_1},\cdots,X_{i_p}\}$. Otherwise, based on Remark \ref{remark:1}, $\Iv_k\subseteq \Iv_{i_1}$. Now, if $\Iv_k\subsetneq \Iv_{i_1}$, then $X_k\in \Sv_{i_1}$ which is a contradiction. Thus, $\Iv_k$ should be equal to $\Iv_i$ but in that case $X_k$ is in the set $\setX_i$. Thus, we can conclude that at least one of the variables in this set is in $\Tv$. Please note that at most one of the variable in $\setX_i$ can be in $\Tv$. Otherwise, the variables in this set cannot have the same indicator set. Therefore, exactly one of the variables in $\setX_i$ is in $\Tv$ and we can obtain the corresponding recovered noise, which is the only recovered noise in $\setN_i$. Without loss of generality, suppose that $X_{i_1}\in \Tv$. Then, based on Remark \ref{remark:1}, other variables in  $\{X_{i_2},\cdots,X_{i_p}\}$ should be descendant of $X_{i_1}$. 
The set $\Sv_{i_1}$ contains the ancestors of $X_{i_1}$. Thus, it also includes parents of $X_{i_1}$. Now, for any path between $N_{i_1}$ and $X_{i_j}$ in the augmented graph $\GT$, $j\geq 2$, if it is outgoing from node $X_{i_1}$, then it is blocked by $X_{i_1}$. If it is in-going toward $X_{i_1}$, then it is blocked by a parent of $X_{i_1}$ which is inside the set $\Sv_{i_1}$. Thus, we can imply that $\tilde{N}_{i_1} \indep X_{i_j} | \{X_{i_1}\}\cup \Sv_{i_1}, \mbox{ for all } 2\leq j \leq p$ under Assumption \ref{assump:faithfulness}. Please note that the variables in $\{X_{i_2},\cdots,X_{i_p}\}$ cannot satisfy the condition in \ref{proposition-3} as they cannot block the path $N_{i_1}\rightarrow X_{i_1}$ by any set $\Sv_{i_j}$ for $j\geq 2$ and the proof is complete.

\subsection{Proof of Theorem \ref{th:main}}
It can be easily seen that for any variable $X_i\in \Xv$, only one of the conditions in \ref{proposition-1}-\ref{proposition-3} in Proposition \ref{th:main_condition} is satisfied. Moreover, these conditions cover all possible cases regarding the relation of the set  $\cup_{j:X_j\in \Sv_i} \Iv_j$ and  $\Iv_i$ and also the uniqueness of $\Iv_i$ in $\mathcal{I}$. Furthermore, in either case, we know definitely whether 
$X_i$ is in $\Tv$ or not. Thus, the intervention target $\Tv$ can be identified uniquely by checking these three conditions for any variable $X_i\in\Xv$ and the proof is complete.

\subsection{Proof of Proposition \ref{th:condition_latent}}


The proof of the three statements in the proposition are as follows.

\ref{proposition-1} The same proof of (I) in Proposition \ref{th:main_condition} in Appendix \ref{app:proof of prop 3} can be applied here.

\ref{proposition-1-L} By contradiction, suppose $X_i\in \TOv$. Then its corresponding exogenous noise $\tilde{N}_i$ must be in $\setN_i$. There are two cases: (1) $\tilde{N}_i$ does not belong to any other indicator set. (2) $\tilde{N}_i$ belongs to another indicator set $\Iv_{j_l}$ for some $X_{j_l}\neq X_i$. Then $X_{j_l}$ must be a descendant of $X_i$, hence $\bI_i\subseteq \bI_{j_l}$ under Assumption \ref{assump:faithfulness}. Therefore both cases contradict the condition in (IV).

\ref{proposition-3-L} 
Similar to the proof of \ref{proposition-3} in Appendix \ref{app:proof of prop 3}, each noise in $\setN_i$ corresponds to either a latent confounder, or an observed variable in $\setX_i$. Further, at most one noise corresponds to a variable in $\setX_i$.

We first present the following remark:
\begin{remark}
Any root variable in $\setX_{i}$ (i.e., variable with no parent in $\setX_{i}$) must be a child of all recovered noises in $\setN_{i}$ in the augmented graph $\GT$. (Recall that in the definition of $\GT$, all latent variables in $\TLv$ are replaced by their corresponding exogenous noise.)
\label{remark:proof-th2-2}
\end{remark}

To prove Remark \ref{remark:proof-th2-2}, suppose $X_{i_k}$ is a root variable in $\setX_i$. Then for any noise $\tilde{N}_l\in \setN_i$, and its corresponding variable $X_l$, there are only two possible cases: (1) $X_l\in \setX_i$, (2) $X_l\in \TLv$, and it is not a parent of any observed variable in $\Sv_i$. For case (1), since $X_{i_k}$ is a root variable, $X_l$ must be $X_{i_k}$, otherwise $X_{i_k}$ has a parent in $\setX_i$. Similarly, for case (2), $X_l$ must be a parent of $X_{i_k}$, hence $N_l$ is a direct parent of $X_{j_k}$.

In the following we prove \ref{proposition-3-L} by contradiction. Suppose $X_{i_j}\in \TOv$. Then all other variables in $\setX_i$ are descendants of $X_{i_j}$, and $X_{i_j}$ is the only root variable in $\setX_i$. Then according to Remark \ref{remark:proof-th2-2}, for each recovered noise $\tilde{N}_l\in \setN_{i}$, there will be an edge from the corresponding noise $N_l$ to $X_{i_j}$ in $\GT$, no matter whether $X_l$ is an observed variable (i.e., $X_l=X_{i_j}$) or a latent confounder. This edge cannot be blocked by any other variables or noises, and it implies that Condition \eqref{eq:third_cond_latent} cannot be satisfied for $X_{i_j}$, which leads to contradiction.

Lastly, we describe why \ref{proposition-2} does not necessarily hold any more. This is because $X_k$ in the proof of \ref{proposition-2} may be a latent variable (i.e., in $\TLv$). In this case, $\bI_k$ is not observable, and $X_k\not\in \Sv_i$. However, if we are given the prior knowledge that $\TLv=\emptyset$, then \ref{proposition-2} still holds.

\subsection{Proof of Theorem \ref{th:main_latent}}
To better distinguish between intervention targets in $\TOv$ and in $\TLv$, we use $\tilde{N}_{h_i}$ to represent the exogenous noise corresponding to $X_i$ for all $X_i\in \TLv$. \new{For ease of notation, we denote the true noises $N_j$ in the auxilary graph as $\tilde{N}_j$ in the following proof, since there is a one-on-one correspondence between them.}


\subsubsection{Proof of Sufficiency} 
We show that if an observed variable $X_i$ is a child of some noise $\tilde{N}_j\in\TNT$ in $\aux$, then it is included in the output $\Kv$ of the LIT algorithm. We consider each of the following four cases:
\begin{enumerate}[(i), leftmargin=*]
\item $X_i\in \Tv$.

\item $X_i\not\in \Tv$, and $\tilde{N}_j$ is the exogenous noise of some observed variable $X_j$.

\item $X_i\not\in \Tv$, $\tilde{N}_j$ is the exogenous noise of some latent confounder $H_j$, and $H_j$ is a parent of $X_i$ in $G$.

\item $X_i\not\in \Tv$, $\tilde{N}_j$ is the exogenous noise of some latent confounder $H_j$, and $H_j$ is not a parent of $X_i$ in $G$.
\end{enumerate}
In the following we show that, under each of these four cases, $X_i$ is included in the output $\Kv$. That is, $X_i$ violates the condition in \ref{proposition-1}, violates the condition in \ref{proposition-1-L}, and either satisfies the condition in \ref{proposition-2} or violates \eqref{eq:third_cond_latent} in \ref{proposition-3-L}. Note that the conditions in \ref{proposition-2} and in \ref{proposition-3-L} only depends on the uniqueness of the indicator set, therefore we do not need to check \ref{proposition-2}.

\textbf{Case (i)}. If $X_i\in \Tv$, then its corresponding exogenous noise $\tilde{N}_i$ satisfies $i \in \tilde{N}_i$. Therefore the condition in \ref{proposition-1} Proposition \ref{th:main_condition} is not satisfied. Further, any observed variable $X_{j_l}$ with $i\in \Iv_{j_l}$ must be a descendant of $X_i$, and hence satisfies $\Iv_i\subseteq \Iv_{j_l}$. Therefore the condition in \ref{proposition-1-L} is not satisfied. Lastly, if there are no other variables that have the same indicator set, then $X_i\in \Kv$ according to \ref{proposition-2}. Otherwise, all other variables that have the same indicator set as $\Iv_i$ must be descendants of $X_i$ (because of $\tilde{N}_i$), hence $X_i$ is a child of any recovered noise in $\setN_i$ (according to Remark \ref{remark:proof-th2-2}). Therefore there does not exist $l$ such that \eqref{eq:third_cond_latent} holds, which means that $X_i\in \Kv$.

\textbf{Case (ii)}. If $X_i\not\in \Tv$ and $\tilde{N}_j$ is the exogenous noise of some observed variable $X_j$, then $\Iv_j=\Iv_i$, and there is at least one inducing path from $\tilde{N}_j$ to $X_i$. This means that any variable $X_k\in \Sv_i$ is not a descendant of $\tilde{N}_j$ in {$G_{\Tv}$}, which implies that $j\in \Iv_i$ but not in $\cup_{j':X_{j'}\in \Sv_{i}}\Iv_{j'}$. Therefore the condition in \ref{proposition-1} is not satisfied. Next, note that $j\in \setN_i=\setN_j$, $\Iv_i=\Iv_j$, and \eqref{eq:first_cond_latent} in \ref{proposition-1-L} does not hold for $X_j$ (explained in Case (i)). Therefore \eqref{eq:first_cond_latent} in \ref{proposition-1-L} does not hold for $X_i$ either. 

Lastly, note that there is at least one other variable ($X_j$) that has the same indicator set as $X_i$, and $X_j$ is a root variable in $\setX_i$ (following the same argument as in Case (i)). Therefore $X_j$ is directly connected to all recovered noises in $\setN_i$. Since $\tilde{N}_j$ is only directly connected to $X_j$, this means that if there is an inducing path from $\tilde{N}_j$ to $X_i$, then for any $l\in \setN_i$, by changing the first edge on this path from $\tilde{N}_j\rightarrow X_j$ to $\tilde{N}_l\rightarrow X_j$, the new path is also an inducing path from $\tilde{N}_l$ to $X_i$. Therefore there does not exist $l$ such that \eqref{eq:third_cond_latent} holds, which means that $X_i\in K$.

\textbf{Case (iii)}. If $X_i\not\in \Tv$, $\tilde{N}_j$ is the exogenous noise of latent confounder $H_j$, and $H_j$ is a parent of $X_i$ in $G$, then according to Definition \ref{def:auxillary_graph}, for all other children $X_k$ of $H_j$, we have $\Iv_i\subseteq \Iv_k$. This immediately implies that the condition in \ref{proposition-1-L} is not satisfied. This also implies that for any variable $X_{k'}\in \Sv_i$, $j\not\in \Iv_{k'}$. Therefore, we have $j\in \Iv_i$ but not in $\cup_{j':X_{j'}\in \Sv_{i}}\Iv_{j'}$, which means that the condition in \ref{proposition-1} is not satisfied. Lastly, if $\Iv_i$ is unique, then $X_i\in \Kv$. Otherwise, since there is an inducing path from all recovered noises in $\setN_i$ to $X_i$, there does not exist $l$ such that \eqref{eq:third_cond_latent} holds. Therefore $X_i\in \Kv$.

\textbf{Case (iv)}. If $X_i\not\in \Tv$, $\tilde{N}_j$ is the exogenous noise of latent confounder $H_j$, and $H_j$ is not a parent of $X_i$ in $G$, then according to Definition \ref{def:auxillary_graph}, there exists some $X_k$ such that $H_j$ is directly connected to $X_k$, and $\Iv_k=\Iv_i$. Further, there is an inducing path from $\tilde{N}_j$ to $X_i$. Note that all children of $\tilde{N}_j$ after part (b) have the same indicator set, which is the same as $\Iv_i$. Therefore the condition in \ref{proposition-1} does not hold. Further, since the condition in \eqref{eq:first_cond_latent} does not hold for $X_k$ (explained in Case (iii)) and $\Iv_k=\Iv_i$, the condition in \eqref{eq:first_cond_latent} does not hold for $X_i$ either. Lastly, since there is an inducing path from all recovered noises in $\setN_i$ to $X_i$, there does not exist $l$ such that \eqref{eq:third_cond_latent} holds. Therefore $X_i\in \Kv$.

\textbf{Conclusion}: We show that if an observed variable $X_i$ belongs to $\cup_{\tilde{N}_i\in \TNT} Ch_{Aux(G_\Tv)}(\tilde{N}_i)$, i.e., it is a child of some noise $\tilde{N}_j\in\TNT$ in $\aux$, then it is included in the output $\Kv$ of the LIT algorithm.

\subsubsection{Proof of Necessity} 
We show that if an observed variable $X_i$ is included in the output $\Kv$ of the LIT algorithm, then it must be a child of some noise $\tilde{N}_j\in\TNT$. That is, if $X_i$ violates the condition in \ref{proposition-1} (i.e., $\setN_i\neq\emptyset$), and the condition in \ref{proposition-1-L}, and either:
\begin{enumerate}[(i), leftmargin=*]
\item $\Iv_i$ is unique, or
\item $\Iv_i$ is not unique, and among all variables with the same indicator set, the condition in \eqref{eq:third_cond_latent} does not hold for $X_i$, 
\end{enumerate}
then it must be a child of some noise $\tilde{N}_j\in\TNT$.

\textbf{Case (i)}. If $\Iv_i$ is unique. This implies that any recovered noise $\tilde{N_j}\in\setN_i$ is a parent of $X_i$ in $\GT$. This is because otherwise the child of $\tilde{N_j}$ is a ancestor of $X_i$ but does not belong to $\Sv_i$. Hence it has the same indicator set as $\Iv_i$, which violates the uniqueness of $\Iv_i$. 

Consider all these recovered noises $\tilde{N_j}\in\setN_i$. If there exists $\tilde{N_j}$ such that $X_i$ is the only child of $\tilde{N_j}$, then $\tilde{N}_j$ is the exogenous noise of $X_i$, i.e., $X_i\in \TOv$, which is a subset of $\cup_{\tilde{N}_i\in \TNT} Ch_{Aux(G_\Tv)}(\tilde{N}_i)$. Otherwise, if all noises $\tilde{N_j}$ has at least two children, then all of them correspond to the exogenous noises of latent confounders. Since the condition in \ref{proposition-1-L} is violated, there exist $l\in \setN_i$ such that for all $j_l$ with $l\in \Iv_{j_l}$, $\Iv_i\subseteq \Iv_{j_l}$. This implies that the indicator set of any other child of $\tilde{N}_l$ in $\GT$ must be a (strict) superset of $X_i$. Therefore the edge $\tilde{N}_l\rightarrow X_i$ is kept according to part (b)(i) of Definition \ref{def:auxillary_graph}, i.e., $X_i$ is the child of $\tilde{N}_l$ in $\aux$.

\textbf{Case (ii)}. If $\Iv_i$ is not unique. This means that $\setX_{i}$ is not a singleton. Consider the root variables in $\setX_{i}$. Note that following the induced causal order induced on $\setX_{i}$, there is at least one root variable. If $X_i$ is a root variable, then according to Remark \ref{remark:proof-th2-2}, all recovered noises in $\setN_i$ is a parent of $X_i$ in $\GT$. In this case we can apply the same proof as in Case (i) to show that $X_i$ is a child of some $\tilde{N}_j$ in $\aux$. That is, if there exists $\tilde{N_j}$ such that $X_i$ is the only child of $\tilde{N_j}$, then $X_i\in \TOv$. Otherwise, if all noises $\tilde{N_j}$ has at least two children, then since the condition in \ref{proposition-1-L} is violated, part (b)(i) in Definition \ref{def:auxillary_graph} is satisfied, and there is an inducing path from all recovered noises in $\setN_i$ to $X_i$ (because a direct connection is an inducing path). Therefore $X_i$ is the child of all $\tilde{N}_j$ in $\aux$.

Next, we consider the case if $X_i$ is not a root variable. If there exists $\tilde{N}_j\in \setN_i$ such that it has only one direct child $X_k$ in $\GT$, then $X_k$ is a root variable in $\setX_{i}$. Since \eqref{eq:third_cond_latent} does not hold for $X_i$, this means that $\tilde{N}_j$ is not independent of $X_i$, conditioned on all observed ancestors of $X_i$ in $\GT$ (i.e., $An_{\GT}(X_i)\setminus (\{X_i\}\cup (\Lv\backslash \Tv))$). This is because all observed ancestors of $X_i$ in $\GT$ must belong to one of the three following cases: observed variable whose indicator set is a strict subset of $\Iv_i$ (i.e., belongs to $\Sv_{i}$), observed variable whose indicator set is the same as $\Iv_i$ (i.e., belongs to $\setX_{i}$), and the recovered exogenous noise of a latent confounder (i.e., belongs to $\cup_{l'\in \Iv_{i_1}\setminus\{j\}}~ \tilde{N}_{l'}$). 

In the following we show that the path that cannot be blocked by these observed ancestors is an inducing path. Suppose the path $\tilde{N}_j\rightarrow X_k \mathdash V_{k_1} \mathdash V_{k_2} \cdots X_i$ is not blocked, where $\{V_{k_m}\}$ are variables in $\GT$, and $\mathdash$ represents either $\leftarrow$ or $\rightarrow$. Note that $X_k$ is an ancestor of $X_i$ thus it is in the condition set. Therefore $X_k$ is a collider on this path, and we have the edge $X_k \leftarrow V_{k_1}$. This means that $V_{k_1}$ is a non-collider on this path and an ancestor of $X_i$ in $\GT$. Since the path is active, $V_{k_1}$ is a latent confounder that is not changing across environments (i.e., belongs to $\Lv\setminus \Tv$). Therefore there is an edge $V_{k_1}\rightarrow V_{k_2}$.
Next consider $V_{k_2}$. If $V_{k_2}$ is a non-collider and is not in the conditioning set (i.e., not an ancestor of $X_i$), then there is an edge $V_{k_2}\rightarrow V_{k_3}$. Since $V_{k_2}$ is not an ancestor of $X_i$, neither is $V_{k_3}$. Therefore $V_{k_3}$ is also a non-collider and we have $V_{k_3}\rightarrow V_{k_4}$. Repeat the same analysis on $V_{k_m}$ for all $m=4,5,\cdots$, the path is $\tilde{N}_j\rightarrow X_k \leftarrow V_{k_1} \rightarrow V_{k_2}\rightarrow V_{k_3} \rightarrow \cdots \rightarrow X_i$. This violates the claim that $V_{k_2}$ is not an ancestor of $X_i$. Therefore, $V_{k_2}$ is a collider and is in the conditioning set (i.e., an ancestor of $X_i$). Then we have the edge $V_{k_2}\leftarrow V_{k_3}$, and following the same analysis in $V_{k_1}$ on $V_{k_3}$, we have $V_{k_3}\in \Lv\setminus \Tv$ and is a confounder. Repeat the same analysis on $V_{k_m}$ for all $m=3, 4, 5,\cdots$, we have that variables $V_{k_m}$ are either confounders in $\Lv\setminus \Tv$, or colliders that are ancestors of $X_i$. Therefore this path is an inducing path from $\tilde{N}_j$ to $X_i$. According to part (a) in Definition \ref{def:auxillary_graph}, $X_i$ is the child of $\tilde{N}_j$ in $\aux$.

Next, we consider the case when each recovered noise $\tilde{N}_j\in \setN_i$ has at least two children in $\GT$. 
Similar to the above argument, there is at least one root variable $X_k$ in $\setX_{i}$ that is an ancestor of $X_i$. This means that $\Iv_k=\Iv_i$, and the edge cannot be removed in part (b)(ii) in Definition \ref{def:auxillary_graph}.
Further, for each recovered noise $\tilde{N}_j\in \setN_i$, since \eqref{eq:third_cond_latent} does not hold for $X_i$, $\tilde{N}_j$ is not independent of $X_i$, conditioned on all observed ancestors of $X_i$ in $\GT$. This means that there is a path from $\tilde{N}_j$ to $X_i$ that is not blocked by the observed ancestors. Similar to the argument above, suppose the path $\tilde{N}_j \rightarrow V_{k_1} \mathdash V_{k_2} \cdots X_i$ is not blocked. If $V_{k_1}$ is a non-collider and is not an observed ancestor of $X_i$ (i.e., not in the conditioning set), then there is an edge $V_{k_1} \rightarrow V_{k_2}$, and $V_{k_2}$ is not an ancestor of $X_i$ either. This means that $V_{k_2}$ is also a non-collider and there is an edge $V_{k_2} \rightarrow V_{k_3}$. Repeat the same analysis on $V_{k_m}$ for all $m=3, 4, 5,\cdots$, the path is $\tilde{N}_j \rightarrow V_{k_1} \rightarrow V_{k_2}\rightarrow V_{k_3} \rightarrow \cdots \rightarrow X_i$. This violates the claim that $V_{k_2}$ is not an ancestor of $X_i$. Therefore, $V_{k_1}$ is a collider and is an ancestor of $X_i$. Then we can repeat the same analysis as above to conclude that this path is an inducing path. Therefore, there is an inducing path from each recovered noise $\tilde{N}_j\in \setN_i$ to $X_i$, hence $X_i$ is a child of $\tilde{N}_j$ in $\aux$, according to part (b)(ii) in Definition \ref{def:auxillary_graph}.

\textbf{Conclusion}: We show that if an observed variable $X_i$ is included in the output $\Kv$ of the LIT algorithm, then it must be a child of some recovered noise in $\aux$, hence belongs to 
$\cup_{\tilde{N}_i\in \TNT} Ch_{Aux(G_\Tv)}(\tilde{N}_i)$.

\subsection{Proof of Remark \ref{remark:comparison}}
Regarding the equivalency between the condition \eqref{eq:third_cond_latent} and the condition \eqref{eq:third_cond} under causal sufficiency assumption, note that under causal sufficiency assumption, the condition in \eqref{eq:third_cond_latent} can be rewritten as follows:
\begin{itemize}[leftmargin=11mm]
\item[(III-L)] Let $\mathcal{X}_i = \{X_{i_1},\cdots,X_{i_p}\}$, for some $p\geq 2$, be a set of all variables whose corresponding indicator sets are the same as $X_i$ and $\setN_i \neq \emptyset$. Then for each $j\in [p]$, $X_{i_j}\not\in \Kv$ if and only if 
the following condition holds:
\begin{equation}
\exists K \subseteq [p]\setminus \{j\},~ \exists\Sv \subseteq \Sv_{i_1},~ \text{ s.t. }
\tilde{N}_l \indep X_{i_{j}} | \left(\cup_{k'\in K} X_{i_{k'}}\right) \cup \Sv.
\tag{C2}
\end{equation}
\end{itemize}

Please note that $\mathcal{N}_i$ only contains $\tilde{N}_l$ (otherwise, at least two variables are descendants of each other which is impossible) under causal sufficiency assumption and each recovered noise only has one children (i.e., their corresponding intervention target) and cannot block any path. Moreover, for any $X_{i_j}$, the set $K=\{i_k\}$ and $\Sv=\Sv_{i_1}$ is enough to guarantee that $\tilde{N}_l \indep X_{i_{j}} | X_{i_k} \cup \Sv_{i_1}$ where $X_{i_k}\in \Tv$ and $\tilde{N}_l$ is its corresponding recovered noise. Thus, all $X_{i_j}$s, $j\neq k$ are excluded from $\Kv$. Moreover, $X_{i_k}$ is added to $\Kv$ as $\tilde{N}_l$ is the direct parent of $X_{i_k}$ and cannot be d-separated by a subset of observed variables and recovered noises.

About comparing our result with \cite{jaber2020causal} In the case that $\Tv_{\Ov}=\Tv$, the augmented graph in \cite{jaber2020causal} is constructed as follows (which we show it by $G'_{\Tv}$). For any pairs of environments such as $E_i$ and $E_j$, a node denoted by $F_{ij}$ is added to the original graph, and it is connected to all variables in $\Tv_{\Ov}$. Please note that in our setting, all variables in $\Tv$ are changing across the environments and all $F_{ij}$s are directly connected to $\Tv_{\Ov}$. The candidate set in \cite{jaber2020causal} is the neighbors of $F$-nodes in the MAG of the augmented graph. Now, based on Theorem \ref{th:main_latent}, it is just needed to show that if some $\tilde{N}_i$ is a parent of some $X_j$ in $\aux$, then some $F$-node is also connected to $X_j$ in MAG of $G'_{\Tv}$. If $X_j$ is $\Tv_{\Ov}$, this is trivial as it should be connected to some recovered noise in $\aux$ and also to some $F$-node in the MAG of $G'_{\Tv}$. Thus, suppose that the $X_i\not\in \Tv_{\Ov}$. In that case, based on part (a)(i) of Definition \ref{def:auxillary_graph}, there should be an inducing path between $\tilde{N}_i$ and $X_j$ relative to $\Lv$ in $G_{\Tv}$. This path starts from $\tilde{N}_i$ and goes directly to its corresponding observed variable in $\Tv_{\Ov}$ (without loss of generality, let us say $X_i$) and continues until getting to $X_j$ in the augmented graph $\GT$. Note that since there are no changing latent variables, this path only involves variables in $\Ov$ and $\Lv$. Therefore we can construct the same path on the original graph $G$. We denote the part of this path in the original graph by $P$. Now, in $G'_{\Tv}$, due to symmetry among $F$-nodes, consider any of them. This node is connected to $X_i$ as $X_i\in \Tv$. Now, the path starting from the $F$-node and then going directly to $X_i$ and continues based on $P$ until reaching $X_j$ is also an inducing path in $G'_{\Tv}$. This shows that our identifiability result is stronger than the one in \cite{jaber2020causal} when $\Tv_{\Ov}=\Tv$, because of part (a)(ii) in Definition \ref{def:auxillary_graph}.

\section{Experiment details}
\label{app:exp}

\subsection{Simulation Settings}
\textbf{Data Generating Mechanism.} 
We evaluated the performance of LIT algorithm on randomly generated synthetic models.\footnote{All numerical simulations are run on the CPU of an Apple Macbook Pro laptop with M1 Pro chip.}
Specifically, we considered each of the three settings below for data generating mechanism:
\begin{enumerate}[(1)]
    \item Linear Gaussian model under causal sufficiency assumption: We randomly generated a directed acyclic graph with $n$ variables, where each edge is connected with probability 1.5/($n-1$). Further,
    we randomly selected  $\lfloor (n+1)/2 \rfloor$ variables to be the intervention targets set $\Tv$, and collected data from $D$ environments. For each variable $X_i\in \Tv$, in the first half of the environment, we sampled the variance of corresponding exogenous Gaussian noise in each environment from the uniform distribution on $[1,5]$. In the second half of the environments, we sampled the variance from the uniform distribution on $[5,9]$. For each variable $X_i\in \Xv\setminus \Tv$, we sampled the corresponding noise variance from the uniform distribution on $[1,3]$ (note that this is the same across all environments). The coefficients (edge weights) in the linear SCM are randomly sampled from the uniform distribution on $[0.5, 1]$. We generated $5000$ number of samples in each environment.
    \item Nonlinear model under causal sufficiency assumption: 
    The generating mechanisms for the causal graph, the intervention targets and the noise variances are the same as Setting (1). We used Laplace noise as the exogenous noises. For each variable $X_i$, the function $f_i(\cdot)$ is a multi-layer perception (MLP) with one hidden layer, where the weights and biases are randomly selected between 0.5 and 1, and the activation function is LeakyReLU with negative slope $k=0.2$. Recall from Appendix \ref{app:recovery} that the data generating model satisfies Assumption \ref{assump:tidleg}(a). 
    We generated 3000 number of samples in each environment.

    \item Linear Gaussian model in the presence of latent confounders: We randomly generated a directed acyclic graph with $n$ variables, where $\lfloor n/2 \rfloor$ variables are latent variables, and each edge is connected with probability 1.5/($n-1$). We ensured that each latent variable has at least two observed children. Further, we randomly select  $\lfloor n/2 \rfloor$ variables to be the intervention target set $\Tv$, where 80\% of the latent variables are in $\Tv$. We used Gaussian noise as the exogenous noises, and the selection of the noise variances are the same as Setting (1). 
    We generated 5000 number of samples in each environment.
    
\end{enumerate}

The number of environments $D$ was set to $\{8, 16, 32\}$ for all three settings.

\textbf{Implementation of LIT Algorithm.} For the recovering phase, in linear models (i.e., Setting (1) and (3)), we used FastICA \cite{hyvarinen2001} to recover the noises. We pruned the entries in the mixing matrix with absolute value lass than 0.2, and consider the support of the pruned mixing matrix as the indicator set. In nonlinear models (i.e., Setting (2)), we ran a contrastive-learning approach similar to TCL \cite{hyvarinen2016unsupervised}, where we trained a four-layer MLP for label classification, and then applied FastICA on the input of the final layer to obtain $\TNT$ (up to certain transformation). Note that we assume that we know the number of non-Gaussian components in all three settings. For recovering the indicator set, for each pair of recovered noise $\tilde{N}_j$ and observed variable $X_i$, we calculate the correlation between $\tilde{N}_j$ and $X_i$ within each environment, and take the sum  over all environments as the overall correlation $\rho_{ij}$. We put $j\in\Iv_i$ if and only if $|\rho_{ij}|\geq 0.2$.

For the matching phase, for causally sufficient models (i.e., Setting (1) and (2))
we use Algorithm 2 to find $\Kv$. Regarding the CI test in \eqref{eq:third_cond}, for each variable $X_{i_k}\in \setX_{i_{[p]}}$, we compute the sum of the partial correlation between $\tilde{N}_l$ and $X_{i_j}$ (conditioning on $X_{i_j}$ and $\Sv_{i_1}$) for all $j\neq k$. We then consider the variable $X_{i_k}$ with the smallest of such sum as the intervention target. For the case with latent confounders (i.e., Setting (3)), we use Algorithm 3 to find $\Kv$. We use a relaxed version of condition \eqref{eq:third_cond_latent}, where we only performed the test for $\Sv=\Sv_{i_1}$ and $K=[p]\setminus \{k\}$ for each $X_{i_k}$. We use partial correlation test with significance level 0.15.

\textbf{Baseline methods.} \new{We compared the performance of LIT algorithm with the following methods: 
\begin{enumerate}
    \item PreDITEr algorithm\footnote{\url{https://github.com/bvarici/uai2022-intervention-estimation-latents}} in \citep{varici2022intervention}. The algorithm returns the same candidate intervention set as in \citep{jaber2020causal} in linear Gaussian SCMs. The algorithm allows for the presence of latent confounders (that are not intervention targets), but assumes the model to be linear Gaussian. 

    \item UT-IGSP algorithm\footnote{\url{https://uhlerlab.github.io/causaldag/utigsp.html}} in \citep{spirtes2000causation}. The algorithm works for both linear and nonlinear SCMs. However, no latent confounders are allowed in the algorithm.

    \item CITE algorithm\footnote{\url{https://github.com/bvarici/intervention-estimation}} in \citep{varici2021scalable}. This is the prior work of PreDITEr algorithm, which only works for linear Gaussian SCMs under causal sufficiency. 
    
    \item FCI-JCI123 algorithm\footnote{\url{https://github.com/caus-am/jci}} in \citep{mooij2020joint}. The algorithm returns a subset of intervention targets in the presence of latent confounders, where no latent variable can be in $\Tv$. Further, the algorithm does not require any assumption on the functional form.
\end{enumerate}
}

We note that FCI-JCI123 algorithm is executable only under the first two settings with $D=8$ due to huge runtime. Therefore we only report FCI-JCI123 algorithm in these two setups. We fine-tuned all three algorithms with different confidence thresholds (parameters), and report the performance of both algorithms with the best performance (PreDITEr with threshold 0.05, UT-IGSP with threshold 0.005, CITE with threshold 0.5, and FCI-JCI123 with threshold 0.2).

There exists some other related work such $\psi$-FCI algorithm in \citep{jaber2020causal} which returns a candidate intervention in the presence of latent confounders or DCDI algorithm in \citep{brouillard2020differentiable} which returns an interventional MEC under causal sufficiency assumption. Regarding $\psi$-FCI algorithm, the current implementation is limited to discrete variables in graphs with a few variables. Moreover, the implementation of DCDI with unknown intervention targets is merely for hard interventions. Thus, we did not compare with these algorithms as their implementations cannot be used in our setting.

\textbf{Metrics.} We compared the recovered intervention targets $\Kv$ with the true intervention targets $\TOv$ based on the following metrics:
\begin{itemize}
    \item Precision: The percentage of the element in $\Kv$ that is also in $\TOv$;
    \item Recall: The percentage of the element in $\TOv$ that is correctly recovered in $\Kv$;
    \item F1 Score: The harmonic mean of Precision and Recall.
\end{itemize}

Additionally, to evaluate how many CI/invariance tests can be reduced by LIT algorithm, we compared the number of CI tests conducted by LIT algorithm, and the number of precision difference estimations (PDE estimates) in PreDITEr algorithm and CITE algorithm, which are used to replace invariance tests in the algorithm \citep{varici2022intervention}. We note that in linear SCMs (i.e., settings 1 and 3), as explained in Appendix \ref{app:efficient alg}, the indicator set $\setI$ can be deduced from the recovered mixing matrix from linear ICA. Therefore we only reported the number of CI tests in \eqref{eq:third_cond} (resp. \eqref{eq:third_cond_latent}) in condition \ref{proposition-3} (resp. \ref{proposition-3-L}) for both settings.

\subsection{Simulation Results}
\begin{figure}[t]
\centering
\includegraphics[width=\textwidth]{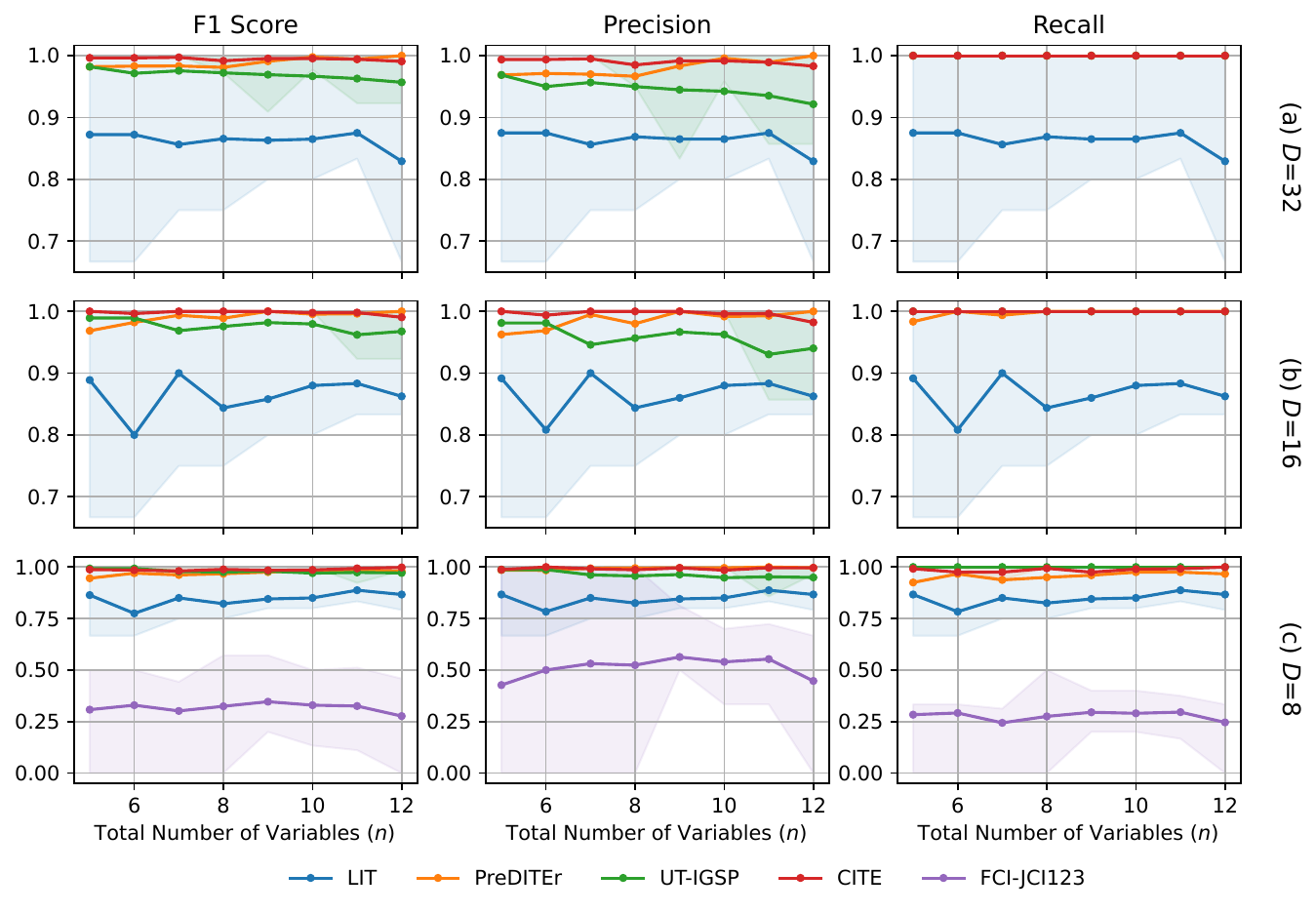}
\caption{Performance of the algorithms under Setting (1) (Linear Gaussian model under causal sufficiency assumption). Error bars represents the 25\% and 75\% percentiles of the corresponding metrics. \new{Note that for $D=32$, PreDITEr algorithm (orange line) overlaps with UT-IGSP algorithm (green line) and CITE algorithm (red line) in recall, as both algorithms have 1.0 recall for all $n$.}
}
\label{fig:setting1} 
\end{figure}

\textbf{Linear Gaussian model under causal sufficiency assumption.}
We selected $n=[5,6,7,8,9,10,$ $11,12]$ and $D=[8, 16, 32]$. We repeated the simulation for 40 times for each $n$ and $D$. \new{We calculated the mean and the errorbar of the metrics, where the errorbar is defined as the 25\% and 75\% percentile among all 40 samples. The plots are shown in Figure \ref{fig:setting1}. The rows represent different number of environments $D$ (32, 24, 8), and the columns represent different metrics (F1 score, precision, recall).}
We observe that PreDITEr and CITE algorithm have the best performance, where the F1 score is almost one. We note that this is because PreDITEr and CITE algorithm are designed specifically for intervention target estimation in linear Gaussian SCMs, which utilizes the sample covariance matrix in the recovery. \new{Besides, UT-IGSP algorithm and LIT algorithm also have decent performances in this setup.} Lastly, we observe that the performance of FCI-JCI123 is significantly worse than the other three methods. We note that the simulation results in \cite{varici2022intervention,mooij2020joint} show that FCI-JCI has comparable performance when the intervention affects the noise distribution to have a mean shift. However, in our setting, the intervention only affects the noise distribution to have a changing variance (scale). 

\begin{figure}[t]
\centering
\includegraphics[width=\textwidth]{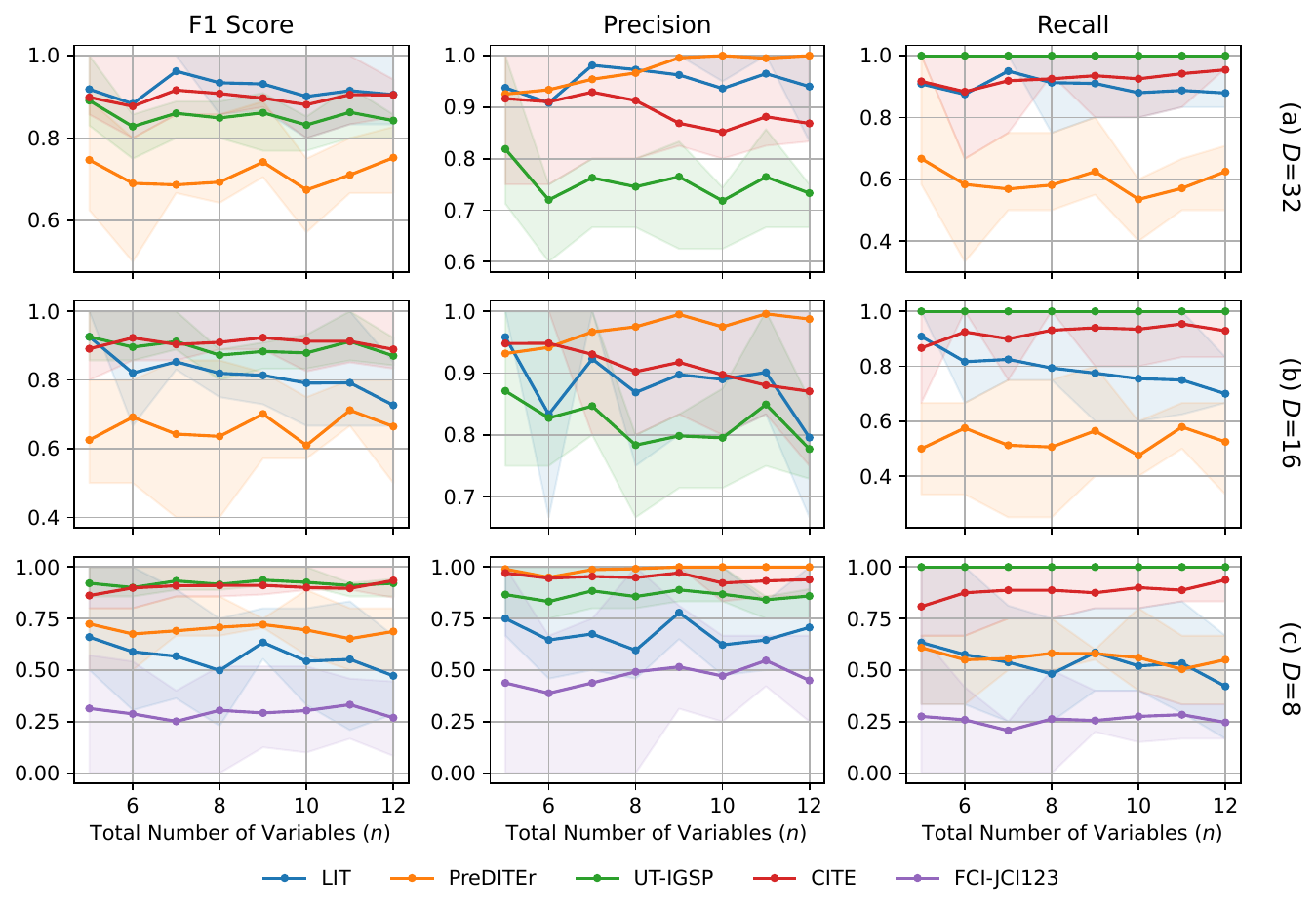}
\caption{Performance of the algorithms under Setting (2) (Nonlinear model under causal sufficiency assumption). Error bars represents the 25\% and 75\% percentiles of the corresponding metrics.
}
\label{fig:setting2} 
\end{figure}

\textbf{Nonlinear model under causal sufficiency assumption.}
We selected $n=[5,6,7,8,9,10,$ $11,12]$ and $D=[8, 16, 32]$, and also repeated the simulation for 40 times for each $n$ and $D$. The average of the metrics are shown in Figure \ref{fig:setting2}.
\new{We observe that the performance of PreDITEr algorithm is worse than in Figure \ref{fig:setting1}. This implies that the performance of PreDITEr algorithm depends heavily on the linear Gaussian assumption of the underlying model. Further, LIT, CITE and UT-IGSP algorithms have similar performance, where LIT performs better when the number of environment is large.} This is because the number of environments has a significant impact on the recovery of the noises in nonlinear ICA. Lastly, FCI-JCI123 still have the worst performance.

\begin{figure}[t]
\centering
\includegraphics[width=\textwidth]{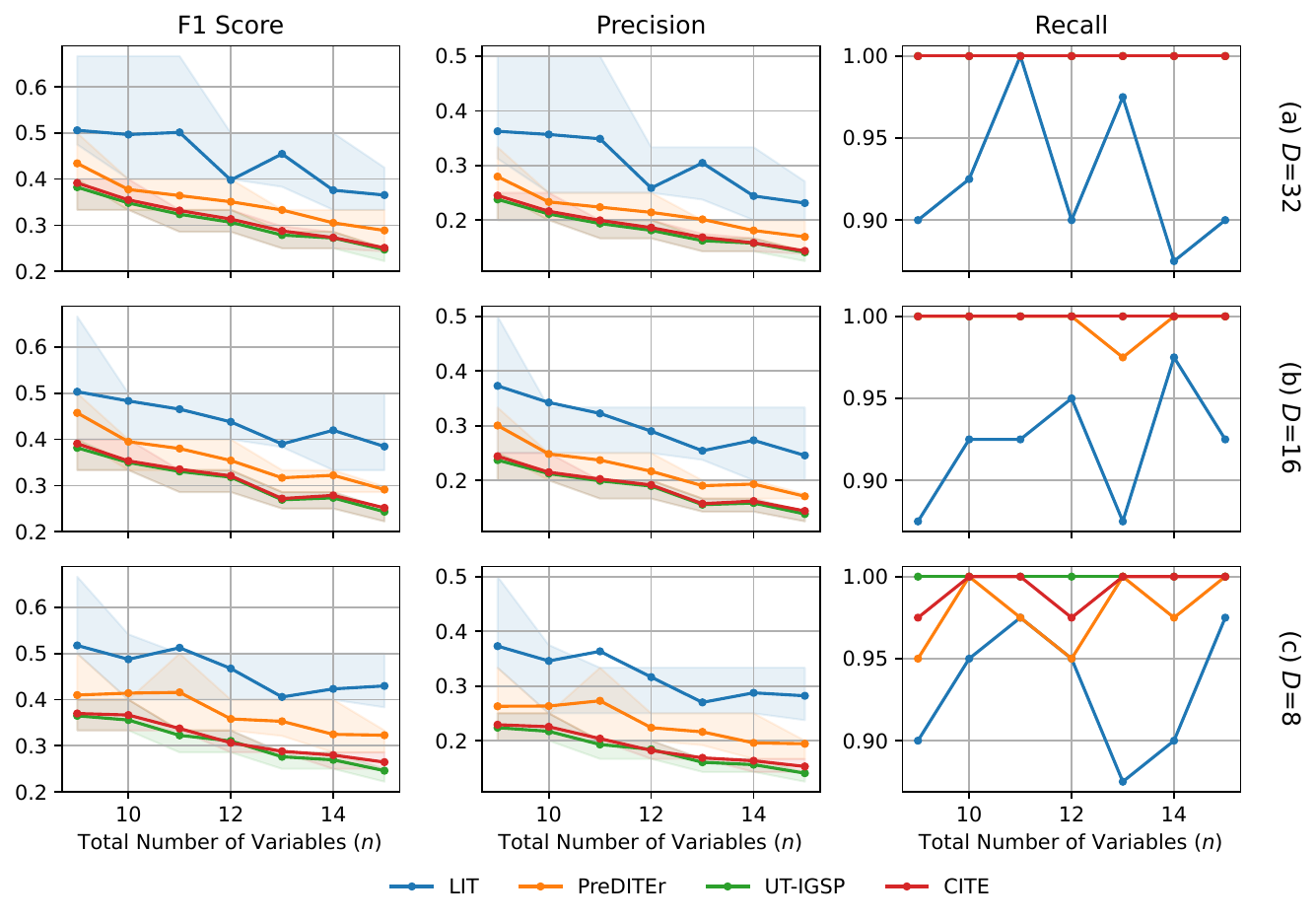}
\caption{Performance of the algorithms under Setting (3) (Linear Gaussian model in the presence of latent confounders). Error bars represents the 25\% and 75\% percentiles of the corresponding metrics.
\new{Note that PreDITEr algorithm (orange line) overlaps with UT-IGSP algorithm (green line) and CITE algorithm (red line) in recall for $D=32$. Further, the are no error bars for recall, because the 25\% percentile is the same as the 75\% percentile (i.e., 1.0) for all three algorithms.}
}
\label{fig:setting3} 
\end{figure}

\textbf{Linear Gaussian model in the presence of latent confounders.}
We selected $n=[9,10,11,12, 13, 14, 15]$ and $D=[8, 16, 32]$. We repeated the simulation for 40 times for each $n$ and $D$, and the average of the metrics are shown in Figure \ref{fig:setting3}. \new{Note that
there are no error bars for recall, because all three high values in recall, and the 25\% percentile is the same as the 75\% percentile (i.e., 1.0) for all three algorithms (i.e., the recall on at least 30 out of 40 randomly generated graphs are one). 

We observe that LIT algorithm has the best performance among all three algorithms. In particular,
all three algorithms have high value in recall, and LIT algorithm has the highest precision. This means that all three algorithms can recover the intervention target in $\TOv$. However, since CITE and UT-IGSP algorithm does not allow for any latent confounders, and PreDITEr algorithm only allows for latent confounders that are not in $\Tv$, all three algorithms cannot distinguish between a changing latent confounder and a changing observed variable.} On the contrary, LIT algorithm can detect the changing latent confounders and thus have a better recovery of the $\TOv$. Lastly, note that in this setting, the average F1-score of both algorithm is much less than one. This is because there are intervention targets that cannot be uniquely recovered, such as the example described in Example \ref{example: latent1}.

\textbf{Number of conducted CI tests.}
\new{We recorded the average number of CI tests conducted by our algorithm, and the average number of PDE estimates by PreDITEr and CITE algorithm. The results for $D=32$ and $D=16$ are shown in Table \ref{tab:1} and \ref{tab:2}, respectively. The results show that LIT algorithm significantly reduces the number of CI tests conducted, especially when $D$ is large. Specifically, the total number of CI tests by LIT algorithm grows quadratically as $n$ increases and is less than 80, while the total number of PDE estimates depends can reach up to 29,600.}

\newpage
\begin{table}[htbp]
  \centering
  \caption{Average number of CI tests in LIT / PDE estimates in PreDITEr and CITE for $D=32$.}
    \begin{tabular}{lcccccccc}
    \toprule
    \textbf{Linear, No latent} & 5     & 6     & 7     & 8     & 9     & 10    & 11    & 12 \\
    \midrule
    LIT   & 2.05  & 2.05  & 1.80  & 2.40  & 2.95  & 2.60  & 2.40  & 3.50 \\
    PreDITEr & 2000.15 & 1931.88 & 3493.40 & 3592.63 & 5486.07 & 5533.10 & 9319.73 & 10404.25 \\
    CITE  & 332.13 & 350.05 & 763.83 & 870.38 & 1713.17 & 1692.85 & 3125.63 & 4139.15 \\
    \midrule
    \midrule
    \textbf{Nonlinear, No latent} & 5     & 6     & 7     & 8     & 9     & 10    & 11    & 12 \\
    \midrule
    LIT   & 16.85 & 21.15 & 30.80 & 35.80 & 48.30 & 54.65 & 70.60 & 76.00 \\
    PreDITEr & 2889.05 & 2753.53 & 5518.70 & 5374.68 & 12155.30 & 12430.90 & 29645.67 & 24483.03 \\
    CITE  & 527.58 & 751.27 & 1542.50 & 1824.65 & 3801.63 & 4569.25 & 11718.85 & 17736.65 \\
    \midrule
    \midrule
    \textbf{Linear, Latent} & 9     & 10    & 11    & 12    & 13    & 14    & 15    &  \\
\cmidrule{1-8}    LIT   & 1.55  & 0.95  & 1.75  & 1.05  & 1.05  & 1.00  & 0.85  &  \\
    PreDITEr & 1583.78 & 1636.88 & 2515.13 & 2541.40 & 3269.72 & 3291.18 & 3016.45 &  \\
    CITE  & 547.77 & 648.23 & 939.23 & 1095.97 & 1665.63 & 1823.78 & 2809.60 &  \\
\cmidrule{1-8}    \end{tabular}%
  \label{tab:1}%
\end{table}%


\begin{table}[htbp]
  \centering
  \caption{Average number of CI tests in LIT / PDE estimates in PreDITEr and CITE for $D=16$}
    \begin{tabular}{lcccccccc}
    \toprule
    {\textbf{Linear, No latent}} & 5     & 6     & 7     & 8     & 9     & 10    & 11    & 12 \\
    \midrule
    LIT   & 1.60  & 2.55  & 1.75  & 2.85  & 2.65  & 2.60  & 1.90  & 2.95 \\
    PreDITEr & 483.85 & 475.75 & 893.65 & 774.38 & 1259.25 & 1392.58 & 2184.47 & 2216.75 \\
    CITE  & 160.38 & 205.93 & 363.48 & 432.77 & 655.30 & 829.27 & 1588.22 & 1735.85 \\
    \midrule
    \midrule
    {\textbf{Nonlinear, No latent}} & 5     & 6     & 7     & 8     & 9     & 10    & 11    & 12 \\
    \midrule
    LIT   & 16.65 & 19.75 & 30.05 & 35.05 & 48.60 & 54.10 & 69.00 & 76.80 \\
    PreDITEr & 667.45 & 693.02 & 1246.72 & 1346.83 & 2804.35 & 2915.47 & 6693.95 & 6379.43 \\
    CITE  & 242.03 & 336.43 & 669.25 & 859.40 & 2015.50 & 2226.18 & 4831.40 & 8201.63 \\
    \midrule
    \midrule
    {\textbf{Linear, Latent}} & 9     & 10    & 11    & 12    & 13    & 14    & 15    &  \\
\cmidrule{1-8}    LIT   & 1.40  & 1.00  & 1.25  & 0.75  & 0.75  & 0.85  & 1.05  &  \\
    PreDITEr & 328.25 & 461.77 & 519.23 & 514.58 & 604.88 & 613.98 & 773.83 &  \\
    CITE  & 265.02 & 324.95 & 468.20 & 463.65 & 947.40 & 916.73 & 1436.65 &  \\
\cmidrule{1-8}    \end{tabular}%
  \label{tab:2}%
\end{table}%

\end{document}

%% file: defns.tex
\newcommand{\Ecal}{\mathcal{E}}

\newcommand{\ev}{{\bf e}}

\newcommand{\gv}{{\bf g}}
\newcommand{\hv}{{\bf h}}

\newcommand{\qv}{{\bf q}}

\newcommand{\vv}{{\bf v}}
\newcommand{\wv}{{\bf w}}

\newcommand{\Av}{{\bf A}}
\newcommand{\Bv}{{\bf B}}
\newcommand{\Cv}{{\bf C}}
\newcommand{\Dv}{{\bf D}}

\newcommand{\Gv}{{\bf G}}

\newcommand{\Iv}{{\bf I}}
\newcommand{\Jv}{{\bf J}}
\newcommand{\Kv}{{\bf K}}
\newcommand{\Lv}{{\bf L}}
\newcommand{\Mv}{{\bf M}}
\newcommand{\Nv}{{\bf N}}
\newcommand{\Ov}{{\bf O}}
\newcommand{\Pv}{{\bf P}}

\newcommand{\Sv}{{\bf S}}
\newcommand{\Tv}{{\bf T}}
\newcommand{\Uv}{{\bf U}}
\newcommand{\Vv}{{\bf V}}
\newcommand{\Wv}{{\bf W}}
\newcommand{\Xv}{{\bf X}}

\newcommand{\Zv}{{\bf Z}}

\def\textiid{i.i.d.\@\xspace}
\newcommand\iid{\ifmmode\text{ i.i.d. } \else \textiid \fi}

\newcommand{\beqs}{\begin{equation*}}
\newcommand{\eeqs}{\end{equation*}}
\newcommand{\beq}{\begin{equation}}
\newcommand{\eeq}{\end{equation}}

\newcommand{\PA}{P\!A}

\newcommand{\TNT}{\tilde{\Nv}_\Tv}
\newcommand{\NT}{\Nv_\Tv}
\newcommand{\GT}{G_\Tv}
\newcommand{\TNTL}{\tilde{\Nv}_{\TLv}}

\newcommand{\setX}{\mathcal{X}}
\newcommand{\aux}{Aux(\GT)}

\newcommand{\mathdash}{\relbar\mkern-9mu\relbar}